\documentclass[onecolumn]{IEEEtran}
 
\usepackage{pgfplots}
\pgfplotsset{compat=1.5}
\pgfplotsset{every axis/.append style={thick}}

\usepackage{hyperref}  
\usepackage{algorithmic}
\usepackage{algorithm} 
\usepackage{booktabs}
\usepackage{amsmath}

\newtheorem{lemma}{Lemma} 
\newtheorem{theorem}{Theorem}  
\newtheorem{proposition}{Proposition} 
\newtheorem{example}{Example}
\newtheorem{definition}{Definition}

\newcommand{\opt}{\mathrm{opt}}
\newcommand{\non}{\mathrm{non}} 
\newcommand*{\defeq }{\stackrel{\text{def}}{=}}  
 
\begin{document}

\title{Average Drift Analysis and Population Scalability}
\author{Jun He and Xin Yao 
\thanks{Manuscript received xx xx xxxx} 
\thanks{This work was supported by the EPSRC under Grant Nos. EP/I009809/1 and EP/I010297/1.   XY was supported by a Royal Society Wolfson Research Merit Award.}
\thanks{Jun He  is  with Department of Computer Science, Aberystwyth University, Aberystwyth, SY23 3DB, UK. Email:    jun.he@aber.ac.uk}
\thanks{Xin Yao is with CERCIA, School of Computer Science, University of Birmingham, Birmingham B15 2TT, UK. Email: x.yao@cs.bham.ac.uk}
}
 
\maketitle

\begin{abstract}     
This paper aims to  study how the population size affects the computation time of evolutionary algorithms in a rigorous way. The computation time of an evolutionary algorithm can be measured by either the expected number of generations (hitting time) or the expected number of fitness evaluations (running time) to find an optimal solution. Population scalability is the ratio of the expected hitting time  between a benchmark algorithm and an algorithm using a larger population size.  Average drift analysis is presented for comparing the expected hitting time  of two algorithms and estimating lower and upper bounds on population scalability.  Several intuitive  beliefs are rigorously analysed. It is prove that (1)  using a population sometimes increases rather than decreases  the expected hitting time; (2) using a population cannot shorten the expected running time of any elitist evolutionary algorithm on unimodal functions in terms of the time-fitness landscape, but this is   not true in terms of the   distance-based fitness landscape; (3) using a population cannot always reduce the expected running time on  fully-deceptive functions, which depends on the benchmark algorithm using elitist selection or random selection.    
\end{abstract}


\section{Introduction}
Population is one of the most important features of evolutionary algorithms (EAs).  A wide range of approaches is available to design population-based EAs. Using a population delivers many benefits~\cite{prugel2010benefits}. The study of the relationship between the performance of an EA and its population size can be traced back to early 1990s. For example, Goldberg et al.~\cite{goldberg1992genetic} presented a population sizing equation to show how a large population size helps an EA to distinguish between good and bad building blocks on some test problems. M\"uhlenbein and Schlierkamp-Voosen~\cite{muhlenbein1993science} studied the critical (minimal) population size that can guarantee the convergence to the optimum.  Arabas
et al.~\cite{arabas1994gavaps} proposed an adaptive scheme for controlling the population size, and the effectiveness of the proposed scheme was validated by an empirical study.  
 Eiben et al.~\cite{eiben1999parameter} reviewed various techniques of parameter controlling for EAs, where the adjustment of population size was considered as an important research issue.  Harik et al.~\cite{harik1999gambler} linked the population size to the quality of solution by the analogy between one-dimensional random walks and EAs.  

The theoretical analysis of the impact of population size on EAs'
computation time starts in early 2000s~\cite{he2002individual}. There has been 
an increasing   interest in rigorously analysing  the relationship 
between the computation time of an EA and its population size. The computation 
time of an EA can be measured by either  the expected
hitting time or the expected  running time. 
The theoretical studies on this topic can be classified into two directions. 

One direction aims to estimate a bound on the computation time of EAs as a function of the population size. This direction belongs to the time complexity analysis of EAs. Drift analysis and tail inequalities are often used for estimating the time bound.  This direction may be called a bound-based study.  A lot of work has done along this direction. The earliest one is  conducted by Jansen et al.~\cite{jansen2005choice} who first obtained the cut-off point for a $(1+\lambda)$ EA on three pseudo-Boolean functions, Leading-Ones, One-Max and Suf-Samp.   J\"agersk\"upper and Witt~\cite{jagerskupper2005rigorous} analysed how the running time of a $(\mu+1)$ EA on the Sphere function scales up with respect to $\mu$.   Witt~\cite{witt2008population} proved theoretically that the running time of a $(\mu+1)$ EA on a specific pseudo-Boolean function is polynomial with an overwhelming probability, when $\mu$ is large enough. Storch~\cite{storch2008choice} presented a rigorous runtime analysis of the choice of the population size with respect to a $(\mu+1)$ EA on several pseudo-Boolean functions. Yu and Zhou~\cite{yu2008new} investigated the expected hitting time  of   $(\lambda+\lambda)$ EAs when $\lambda=1$ and $\lambda =n$ the problem input size  on the trap problem.  Oliveto et al.~\cite{oliveto2008analysis} presented a runtime analysis of both $(1+\lambda)$ and $(\mu+1)$ EAs on some instances of the vertex covering problem.  Friedrich et al.~\cite{friedrich2009analysis} analysed the running time of a $(\mu+1)$ EA  with diversity-preserving mechanisms on the Two-Max problem. 
Chen et al.~\cite{chen2009new} obtained an upper bound on the hitting time of   $(\lambda+\lambda)$ EAs  on Leading-Ones and One-Max problems. L\"assig and Sudholt~\cite{lassig2011adaptive} presented a running time analysis of a $(1+\lambda)$ EA with an adaptive offspring size $\lambda$ on several pseudo-Boolean functions.
Rowe and Sudholt~\cite{rowe2012choice} discussed the running time of $(1+\lambda)$ EAs  in terms of  the offspring population size  on unimodal functions.  
Doerr and
K\"unnemann~\cite{doerr2013how} analysed  the time bound of $(1+\lambda)$ EAs for optimizing linear pseudo-Boolean functions.  Doerr and
K\"unnemann~\cite{doerr2013royal}  showed that $(1+\lambda)$ EAs   with even very large offspring populations does not reduce the runtime significantly on the Royal Road function. Oliveto and Witt~\cite{oliveto2014runtime}  presented a rigorous running time analysis of the well-known Simple Genetic Algorithm  for One-Max.   Gie{\ss}en and Witt~\cite{giessen2015population} studied the relation between the population size and mutation strength for a $(1+\lambda)$ EA on One-Max.

Another direction aims to calculate a ratio, named population scalability, which is described as follows:
\begin{align}
\label{equScalability} \frac{\text{expected hitting time of a benchmark EA}}{ 
\text{expected hitting time of an EA using a population}}.
\end{align} This direction may be called a ratio-based study. 
As one of the earliest analyses, He and Yao~\cite{he2002individual} investigated how the population scalability of   EAs varies as the population size changes on two simple functions (One-Max and Deceptive). In that 
paper, EAs are assumed to be run on a hypothetical parallel computer, that is, 
to assign each individual on one processor. If  the communication
cost is ignored, the population scalability is equivalent to the speedup in  parallel 
computation. The link between scalability and parallelism was further discussed
 in \cite{he2006analysis}. However, since it is too hard to calculate the ratio (\ref{equScalability}), no further development of population scalability has been 
made since then.

This paper belongs to the ratio-based study. It is significantly difference from the bound-based study. The bound-based study focuses on an asymptotic  time bound. It does not intend to calculate the ratio (\ref{equScalability}). The ratio-based study  aims to calculate the ratio (\ref{equScalability}) and it is not necessary to estimate the time bound. 

Compared with previous work on the analysis of population-based EAs~\cite{he2002individual}, the current paper has two novelties. First,  average drift analysis is presented as a  tool of comparing the expected hitting time  of two EAs and  studying population scalability.  The approach used in~\cite{he2002individual} is based on the fundamental matrix of an absorbing Markov chain. It is hard to calculate the expected hitting time through the fundamental matrix. But with average drift analysis, it is possible to estimate the population ratio without calculating the expected  hitting time. 

Secondly,  
the scalability threshold replaces the cut-off point.    The population threshold is the minimal population size at which the running time of an EA using a larger population size is greater than that of the benchmark EA. The cut-off point \cite{jansen2005choice} is the maximize population size at which the running time of the EA is the same order as that of the benchmark EA. 
Let's show the advantage of population scalability by an example: $(1+\lambda)$ EAs (using bitwise mutation and elitist selection) for solving the One-Max problem. According to   \cite{jansen2005choice}, the cut-off point  is
$
 \Theta\mathord{\left(\frac{(\ln n)(\ln
\ln n) }{\ln \ln \ln n}\right)}.
$
This means when the population size $\lambda$ is smaller than the  cut-off point,
the running time  of the $(1+\lambda)$ EA   is  in the same order as that of the $(1+1)$ EA but different  by a constant factor. The constant could be $1000$ or $1/1000$. Therefore the cut-off point does not answer the question whether the expected running time of a $(1+\lambda)$ EA (where $\lambda \ge 2$) is smaller or larger than that of the $(1+1)$ EA. However,  according to   Proposition~\ref{proUnimodal3} and its discussion in this paper, the scalability threshold   is $2$. This means that the running time of the $(1+\lambda)$ EA (where $\lambda \ge 2$) is   larger than that of the $(1+1)$ EA. 

The paper is organised as follows:  Section~\ref{secEAs} defines population scalability.  Section~\ref{secDriftAnalysis} presents drift analysis for population scalability. Section~\ref{secAnalysis1} analyses scenario 1: using a population does not reduce the hitting time.  Section~\ref{secAnalysis2} analyses scenario 2: using a population reduces the hitting time, but  not the  running time. Section~\ref{secAnalysis3} investigates scenario 3: using a population reduces the running time.  Section~\ref{secConclusions} concludes the paper.

\section{Population scalability}
\label{secEAs}

\subsection{Evolutionary algorithms}

Consider the problem of maximizing a function $f(x)$ where $ x \in \mathcal{S}$ and $\mathcal{S}$ is a finite set. A point in $\mathcal{S}$ is called a solution or  an individual. A population consists of one or more individuals. A $(\mu+\lambda)$  EA  is described in Algorithm~\ref{alg1}. 
The stopping criterion is that  the EA halts once an optimal solution is found. The criterion is used for the sake of   analysis because our interest is the number of generations of the EA to find an optimal solution for the first time. If $\Phi_t $ includes an optimal solution,  assign
$ \Phi_t =    \Phi_{t+1}= \Phi_{t+2} =\cdots$     for ever.  In other words, the optimal set is always absorbing.   
 
\begin{algorithm}
\caption{A $(\mu+\lambda)$ EA where $\mu,\lambda \ge 1$}  \label{alg1}
\begin{algorithmic}[1]
\STATE initialise  a population $\Phi_0$ consisting of $\mu$  individuals (solutions)   and  $t \leftarrow 0$;  
\STATE evaluate the fitness of individuals in $\Phi_0$; 
\WHILE{$\Phi_t$ does not include an optimal solution}
\STATE mutate (or crossover) individuals in $\Phi_t$ and generate a children population $\Psi_t$ consisting of $\lambda$  individuals;
\STATE evaluate the fitness of individuals in $\Psi_t$; 
\STATE probabilistically select $\mu$ individuals from  the   parent population $\Phi_t$ and children population $\Psi_t$, and  form  a new parent population $\Phi_{t+1}$;
\STATE   $t\leftarrow t+1$;  
\ENDWHILE
\end{algorithmic}
\end{algorithm}

The sequence $\{ \Phi_t; t=0, 1, \cdots \}$ can be modelled by a  Markov chain~\cite{he2003towards}. Each generation of the   EA consists of two steps: to generate new individuals by mutation or crossover and to select individuals for next generation,   
\begin{align*}
\Phi_t \overset{\text{mutation (or crossover)}}\longrightarrow \Psi_{t} \cup \Phi_{t}  \overset{\text{selection}}\longrightarrow \Phi_{t+1}.
\end{align*}

Let  $\mathcal{P}$ denote the set of all populations, $\mathcal{P}_{\opt}$ the set of   populations including  an optimal solution and  $\mathcal{P}_{\non}$ the set of   populations without an optimal solution. The  transition from   $\Phi_t$ to  $\Psi_{t}$ can be represented using mutation (or crossover) probabilities:   
\begin{align}
P_{\mathrm{m}}(X,Y)\defeq P(\Psi_{t}=Y \mid \Phi_t =X),  \quad X, Y  \in \mathcal{P},
\end{align}  
where   $\Phi_{t}, \Psi_{t}$ are random variables representing the $t$th generation population and its children population. $X, Y $ are their  values taken from $\mathcal{P}$.  

The  transition from    $\Phi_t$  and $\Psi_{t}$ to  $\Phi_{t+1}$   can be represented using selection probabilities:
\begin{align}
 P_{\mathrm{s}}(X,Y,Z)\defeq &P(\Phi_{t+1}=Z \mid \Phi_t =X, \Psi_{t}=Y).  
\end{align}  

The  transition from   $\Phi_{t}$ from $\Phi_{t+1}$ can be  represented using   transition probabilities:  
\begin{align}
P(X,Y)\defeq  \Pr (\Phi_{t+1}= Y \mid \Phi_t=X).
\end{align} 

   The hitting time is the number of generations of an EA to find an optimal solution for the first time. 
   \begin{definition} Given $\Phi_0=X$, the expected hitting time of an $(\mu+\lambda)$ EA is defined by
   \begin{align}
g(X)\defeq  
 & \sum^{+\infty}_{t=0}    \Pr (\Phi_t \in \mathcal{P}_{\non}  )
.
\end{align}
  If $\Phi_0$ is chosen according  to  a probability distribution over $\mathcal{P}$,  the expected hitting time is given by
$$
g(\Phi_0) \defeq  \sum_{X \in \mathcal{P}} g(X)    \Pr (\Phi_0 =X). 
$$ The expected running time of a $(\mu+\lambda)$ EA is the expected number of fitness evaluations, which equals  to $\mu+\lambda g(\Phi_0) $. For the sake of simplicity, we always omit the first term $\mu$, which is the number of fitness evaluations in initialization. 
\end{definition}

If genetic operators don't change as time, the   sequence $\{ \Phi_t; t=0, 1, \cdots \}$ can be modelled by a  homogeneous   Markov chain. According to the fundamental matrix theorem~\cite[Theorem 11.5]{grinstead1997introduction}, the expected hitting time  of an EA can be calculated from transition probabilities.

\begin{theorem}
\label{theFundamentalMatrix}
If population sequence $\{ \Phi_t, t=0,1, \cdots \}$ is a homogeneous  Markov chain and  converges to $\mathcal{P}_{\opt}$, that is, $\lim_{t \to +\infty}    \Pr (  \Phi_{t} \in \mathcal{P}_{\opt}) =1,$   then the expected hitting time $g(X)$ satisfies a linear equation system: 
\begin{equation}
\label{equLinearSystem}
\left\{
\begin{array}{lll}
 g(X) =0, &\scriptstyle \text{if } X\in \mathcal{P}_{\opt},\\
  \sum_{Y \in \mathcal{P}} P (X,Y) \left( g(X) - g(Y)\right) =1,
        &\scriptstyle \text{if } X \notin \mathcal{P}_{\non}.
\end{array}
\right.
\end{equation} 
\end{theorem}
 
The fundamental matrix theorem is useful in analysing elitist EAs~\cite{he2002individual,he2003towards,zhou2007runtime}. However, its disadvantage is the difficulty of solving the linear equation system~(\ref{equLinearSystem}).

\subsection{Population scalability}
 \label{secScalability}

The population scalability is defined as the ratio of the expected hitting time between a benchmark EA and an EA with a larger population size. In this paper, the benchmark is a $(1+1)$  EA using mutation and selection operators.  
Other types of EAs may play the role of a benchmark too. For example, a $(2+2)$ EA could be chosen as a benchmark  when studying  EAs with crossover. But we will not discuss them here.
\begin{definition} 
Given a $(1+1)$  EA and a $(\mu+\lambda)$   EA   that exploit an identical mutation operator to optimise the same fitness function, let  $\Phi_0^{(1+1)}$ and $\Phi_0^{(\mu+\lambda)}$ denote their corresponding initial  populations, then the population scalability  is  defined by
  \begin{align}
   &PS(\mu+\lambda \mid  \Phi_0^{(1+1)}, \Phi_0^{(\mu+\lambda)}) \defeq  \frac{g^{(1+1)}(\Phi^{(1+1)}_0)}{g^{(\mu+\lambda)}(\Phi^{(\mu+\lambda)}_0)},
\end{align} 
where the superscripts $^{(1+1)}$ and $^{(\mu+\lambda)}$ are used to distinguish the $(1+1)$  EA and $(\mu+\lambda)$   EA.  
\end{definition}
 
An essential part of the definition above is that both EAs must adopt identical mutation operators. This ensures that the comparison is meaningful.
Nonetheless, it is impossible for the selection operators to be identical. Indeed even if the selection operators are of the same type, for example roulette wheel selection, the conditional probabilities determining the actual selection operators are never identical under distinct population sizes.

Obviously the value of population scalability relies on initial populations. Due to the use of a population, $\Phi^{(\mu+\lambda)}_0$ may contain several individuals some of which are different from   $\Phi^{(1+1)}_0$. For the sake of comparison, we restrict our discussion to identical initialization, that is,  
 for the $(1+1)$ EA,  $\Phi_0^{(1+1)}=x$ 
and for the $(\mu+\lambda)$ EA,  $\Phi_0^{(\mu+\lambda)}=(x, \cdots, x)$. 
In this case, 
$PS(\mu+\lambda \mid  \Phi_0^{(1+1)}, \Phi_0^{(\mu+\lambda)})$ is denoted by
$PS(\mu+\lambda \mid  x)$ in short. There exist other types of initialization but we will not discussed them here.

The notion of population scalability is similar to that of the speedup widely used when analyzing parallel algorithms. The speedup of parallel EAs have been studied through experiments~\cite{andre1998parallel,alba2002parallel,alba2002parallelism}. If each individual is assigned to a processor, then EAs turn into parallel EAs. Under this circumstance,    population scalability is equivalent to speedup  if ignoring the communication cost. Hence  population scalability is  called  speed-up 
on a hypothetical parallel computer in~\cite{he2002individual}.

The following questions are essential when studying population scalability.  
\begin{enumerate}
\item Given a $\lambda\ge 2$ or $\mu \ge 2$,  is  population scalability
$
PS(\mu+\lambda \mid x)>1?
$

If it is, we may assign  each individual to a processor in a parallel computing system and then the CPU computation time of the $(\mu+\lambda)$ EA  is less than that of the $(1+1)$ EA (if ignoring the communication cost).

\item Given a $\lambda\ge 2$ or $\mu \ge 2$,  is   population scalability
$
PS(\mu+\lambda \mid x)>\lambda?
$
 
 If it is, then  the CPU computation time of the $(\mu+\lambda)$ EA on a computer is less than that of the $(1+1)$ EA.

\item Where are the smallest population sizes $(\mu, \lambda)$ such that the expected running time of the $(\mu+\lambda)$ EA is larger than that of the $(1+1)$ EA?

 We call this point the scalability threshold, which satisfies 
\begin{align}\left\{
\begin{array}{llll}
\min \{\mu: PS(\mu+\lambda \mid x) >\lambda\},\\
\min \{\lambda: PS(\mu+\lambda \mid x) >\lambda\}.
\end{array}
\right.
\end{align} 

In general, the scalability threshold is not a single point but a Pareto front due to minimizing both population sizes $\mu$ and $\lambda$ simultaneously. However, in a $(1+\lambda)$    or $(\lambda+\lambda)$ EA, the scalability threshold is a single point.

\end{enumerate}

\section{Average drift analysis and time-fitness landscape}
\label{secDriftAnalysis}

\subsection{Average drift analysis} 
Average drift analysis is a variant of drift analysis for estimating the expected hitting time of EAs. The idea of average drift was first used by J\"{a}gersk\"{u}pper who     considered average drift  of a $(1+1)$   EA on linear functions and provided a delicate analysis of the running time of the $(1+1)$ EA \cite{jagerskupper2008blend}. Nevertheless the discussion in \cite{jagerskupper2008blend}  was restricted to the $(1+1)$ EA and linear functions. Even the term of average drift did not appear in \cite{jagerskupper2008blend}  but  was first adopted by Doerr~\cite{doerr2011drift} for introducing J\"{a}gersk\"{u}pper's  work  \cite{jagerskupper2008blend}. In this section, the average drift is formally  defined    and  then general average drift theorems are  presented.

In drift analysis,  a distance function $d(X)$ is used to measure how  far a population  $X$ is away from the optimal set $\mathcal{P}_{\opt}$. It  is a non-negative function such that $0<d(X)<+\infty$ for any $X \in \mathcal{P}_{\non}$ and $d(X)=0$ for $X \in \mathcal{P}_{\opt}$.  Drift is used to measure the  progress rate  of a population moving towards the optimal set  per generation.   
\begin{definition}
Given a population  $X$,   the  pointwise drift at $X$ is  
 \begin{align} 
 \Delta (X) \defeq   
 \sum_{Y \in \mathcal{P} } \left( d(X)- d(Y)\right)   \Pr (\Phi_{t+1}= Y\mid  \Phi_t=X). 
\end{align} 
Given a generation $ t$,   the average drift  at $t$ is   
\begin{align}
\label{equAverageDrift} 
 \bar{\Delta}_t   &\defeq
 \left\{
 \begin{array}{llll}
 0, &\scriptstyle \textrm{if }  \Pr (\Phi_t \in \mathcal{P}_{\non})=0,\\       
 \scriptstyle    \frac{    \sum_{X \in \mathcal{P}_{\non} } \Delta (  X)
     \Pr ( \Phi_t=X) }{    \Pr (\Phi_t \in \mathcal{P}_{\non})}, &\scriptstyle \textrm{otherwise}.        
 \end{array}
 \right. 
\end{align} 
\end{definition} 

Since EAs are randomised search, we  make a reasonable assumption in the whole paper:  if $\Pr(\Phi_0 \in \mathcal{P}_{\non})>0$, then  for any fixed generation $t$, $\Pr(\Phi_t \in \mathcal{P}_{\non})>0$. 
 
The following   theorem  provides an approach to estimating  a lower bound  on the expected hitting time. It is  a variation  of \cite[Theorem 4]{he2001drift}.    
\begin{theorem}  
 \label{theDriftLowerBound}  Assume that population sequence $\{ \Phi_t, t=0,1, \cdots \}$ converges to $\mathcal{P}_{\opt}$ where   $\Phi_0$ satisfies $\Pr(\Phi_0 \in \mathcal{P}_{\non})>0$. Given a distance function $d(X)$, if for any $t$,  the average drift $\bar{\Delta}_t  \le c$  where $c>0$,   then    the expected hitting time $  g(\Phi_0) \ge 
d(\Phi_0)/c  
$, where 
$$d(\Phi_0)\defeq \sum_{X \in \mathcal{P}} d(X)    \Pr (\Phi_0 =X).$$ Furthermore if for at least one $t$,  the average drift $\bar{\Delta}_t  < c$,   then   $  g(\Phi_0) >
d(\Phi_0)/c  
$.
\end{theorem}

\begin{IEEEproof}
Without loss of generality,  let $c=1$. From the condition $\bar{\Delta}_t   \le 1$, we have  
   \begin{align}
   & \Pr (\Phi_t \in \mathcal{P}_{\non} ) \nonumber \\
   \ge& \sum_{X \in \mathcal{P}_{\non} } \Delta (  X)
     \Pr ( \Phi_t=X) \nonumber\\ 
 \ge&\scriptstyle   \sum_{X \in \mathcal{P}_{\non}} d(X)     \Pr ( \Phi_t=X)   -    \sum_{Y \in \mathcal{P}_{\non}} d(Y)     \Pr (  \Phi_{t+1}=Y). \label{equProb1}
\end{align}  
Summing  the term  $   \Pr (\Phi_t \in \mathcal{P}_{\non}  )$ from $t=0$ to $k$, we get 
\begin{align}
 & \sum^k_{t=0}    \Pr (\Phi_t \in \mathcal{P}_{\non}  )  \nonumber \\
 \ge &\scriptstyle\sum^k_{t=0}  (    \sum_{X \in \mathcal{P}_{\non}} d(X)     \Pr ( \Phi_t=X)  -   \sum_{Y \in \mathcal{P}_{\non}} d(Y)     \Pr ( \Phi_{t+1}=Y) )  \nonumber \\
    =& \scriptstyle\sum_{X \in \mathcal{P}_{\non}} d(X)    \Pr (\Phi_0=X)   
    -   \sum_{Y \in \mathcal{P}_{\non}} d(Y)      \Pr (  \Phi_{k+1} =Y). \label{equSum1}
\end{align} 
Notice that
\begin{align}
&\textstyle\sum_{Y \in \mathcal{P}_{\non}} d(Y)      \Pr (  \Phi_{k+1} =Y) \nonumber
\\
\le & \textstyle\max_{X \in \mathcal{P}} d(X) \sum_{Y \in \mathcal{P}_{\non}}     \Pr (  \Phi_{k+1} =Y) \nonumber  \\
= &\textstyle\max_{X \in \mathcal{P}} d(X)    \Pr (  \Phi_{k+1}  \in \mathcal{P}_{\non}) . \label{equSum2}
\end{align}
Since the EA is convergent:  $
\lim_{k \to +\infty}    \Pr (  \Phi_{k+1} \in \mathcal{P}_{\non}) =0,
$ then from (\ref{equSum2})  we have
\begin{align}
\lim_{k \to +\infty}  \sum_{Y \in \mathcal{P}_{\non}} d(Y)      \Pr (  \Phi_{k+1} =Y) =0.
\end{align}
Applying the above result  to (\ref{equSum1}) (let $k \to +\infty$), we get
\begin{align}
 g(\Phi_0)=&\sum^{+\infty}_{t=0}    \Pr (\Phi_t \in \mathcal{P}_{\non}  )  \nonumber  \\
 \ge & \sum_{X \in \mathcal{P}_{\non}} d(X)    \Pr (\Phi_0=X)=d(\Phi_0), \label{equSum3}
\end{align}
which  gives the desired result.
 If for some $t$,  the average drift $\bar{\Delta}_t  < 1$,  then inequality (\ref{equSum1}) is strict and inequality (\ref{equSum3}) is strict too.  
\end{IEEEproof}  

Similarly, the theorem below provides an approach to estimating an upper bound on the expected hitting time. It is  a variation of \cite[Theorem 1]{he2001drift}.      Its proof is similar to that of the above theorem. 
\begin{theorem}  
\label{theDriftUpperBound}   
Assume that population sequence $\{ \Phi_t, t=0,1, \cdots \}$ converges to $\mathcal{P}_{\opt}$ where   $\Phi_0$ satisfies $\Pr(\Phi_0 \in \mathcal{P}_{\non})>0$. Given a distance function $d(X)$,
if  for any $t$, the average  drift   $\bar{\Delta}_t    \ge c$ where $c>0$,   then    the expected hitting time  $  g(\Phi_0)\le 
d(\Phi_0)/c$. Furthermore if for at least one $t$,  the average drift $\bar{\Delta}_t  > c$,   then   $  g(\Phi_0) <
d(\Phi_0)/c  
$.
\end{theorem}

The average drift analysis provides a useful tool for comparing the expected hitting time  of two EAs. Its idea is simple. One EA is taken as the benchmark and its expected hitting time is used to set a distance function for the other EA. The average drift of the other EA is estimated and then its expected hitting time  is bounded using  average drift analysis.

\begin{theorem}
\label{theCompareLower}
Given two EAs $A$ and $B$ to optimise the same fitness function, let $ X_0$ and $ Y_0$ be  their initial populations respectively (non-optimal).  
For  algorithm $B$,  define a distance function   $d^{B}(X)$ such that
$
 d^{B}(Y_0)=   g^{A}(X_0),
$ where $g^{A}(X_0)$  is the expected hitting time of  algorithm $A$ starting at $X_0$.
If for any $t \ge 0$,  average drift 
$ 
\bar{\Delta}^{B}_t \le c$  where $c >0$, then   the expected hitting time  of algorithm $B$ satisfies that   $  g^{B}(Y_0) \ge g^{A}(X_0)/c$. Furthermore if for at least one $t \ge 0$,    $ 
\bar{\Delta}^{B}_t < c$, then     $  g^{B}(Y_0) > g^{A}(X_0)/c$.    
\end{theorem}
 
 \begin{IEEEproof}
It is a direct corollary of Theorem~\ref{theDriftLowerBound}.
\end{IEEEproof}

\begin{theorem}
\label{theCompareUpper}
Given two EAs $A$ and $B$ to optimise the same fitness function, let $ X_0$ and $ Y_0$ be  their initial populations (non-optimal) respectively.     
For  algorithm $B$,  define a distance function   $d^{B}(Y)$ such that
$
 d^{B}(Y_0)=   g^{A}(X_0),
$ where $g^{A}(X_0)$  is the expected hitting time of  algorithm $A$ starting at $X_0$.
If  for any $t \ge 0$,    average drift 
$ 
\bar{\Delta}^{B}_t   \ge c$  where $c >0$, then     $  g^{B}(Y_0)\le g^{A}(X_0)/c$. Furthermore if for some $t \ge 0$,    $ 
\bar{\Delta}^{B}_t   > c$, then     $  g^{B}(Y_0) < g^{A}(X_0)/c$.    
\end{theorem}
 
  \begin{IEEEproof}
It is a direct corollary of Theorem~\ref{theDriftUpperBound}.
\end{IEEEproof}

Pointwise drift theorems  are corollaries of average drift theorems, because it requires stronger condition on the pointwise drift:  $\Delta ( X)\ge c$ (or $\le c $) for any $X \in \mathcal{P}_{\non}$. It implies the average drift $\bar{\Delta}_t  \ge   c  $ (or $\le c$) for any $\Phi_0 \in \mathcal{P}_{\non}$.       
\begin{theorem}\cite[Theorem 2]{he2003towards}  
\label{thePointwiseDriftLower} Assume that population sequence $\{ \Phi_t, t=0,1, \cdots \}$     converges to $\mathcal{P}_{\opt}$. Given a distance function $d(X)$, if  for any $  X \in \mathcal{P}_{\non}$, the pointwise drift  $ \Delta ( X)\le c$ (where $ c>0$),  then  for any initial population $X_0 \in \mathcal{P}_{\non}$,   the expected hitting time  $  g(X_0) \ge d(X_0)/c $.
\end{theorem}    

\begin{theorem}\cite[Theorems  3]{he2003towards} 
\label{thePointwiseDriftUpper} Assume that  population sequence  $\{ \Phi_t, t=0,1, \cdots \}$   converges to $\mathcal{P}_{\opt}$. Given a distance function $d(X)$, if   for any $  X \in \mathcal{P}_{\non}$,   the pointwise drift   $ \Delta (  X)\ge c$ where $c>0$,   then  for any initial population $X_0 \in \mathcal{P}_{\non}$,  the expected hitting time   $  g(X_0) \le d(X_0)/c  
$.
\end{theorem} 

\subsection{Average drift analysis for population scalability}
Drift analysis for population scalability is based on an  simple idea.  Given a benchmark EA and another EA, assume that  both EAs start at the same point  with an equal distance to the optimal set. If at each generation, the  progress rate   of the other EA is $10$ times  that of the benchmark EA,  then the expected hitting time of the other EA will be   $1/10$ of that of the benchmark EA. Thus the population scalability is  $10$. This simple idea can be formalised by average drift analysis.  

Consider a ($1+1)$ EA and a $(\mu+\lambda)$ EA (where $\lambda \ge 2  $) that exploit an identical mutation operator to optimise the same fitness function. Assume  that $\Phi^{(1+1)}_0=x_0$ and $\Phi^{(\mu+\lambda)}_0=(x_0, \cdots, x_0)$  for some $x_0 \in \mathcal{P}_{\non}$. 
For the $(\mu+\lambda)$  EA,  define a distance function   $d(X)$ such that
$
 d^{(\mu+\lambda)}(x_0, \cdots, x_0)=   g^{(1+1)}(x_0).
$  

The first  theorem establishes a sufficient condition     for estimating   the upper bound on population scalability. 
Thanks to average drift analysis, there is no requirement that the
population sequence  is a Markov chain.
\begin{theorem} 
\label{thePSUpperBound}
Given a non-optimal  $x_0$, if  for all $t \ge 0$, the average drift 
\begin{align}
\label{equPSUpperBound}
\bar{\Delta}^{(\mu+\lambda)}_t    \le c
\end{align}
where $c>0$, then $PS(\mu+\lambda \mid x_0) \le c.$ Furthermore if for at least one $t \ge 0$,  inequality (\ref{equPSUpperBound}) is strict,  then $PS(\mu+\lambda \mid x_0)<c$. 
\end{theorem}

\begin{IEEEproof}
According to Theorem~\ref{theCompareLower}, the expected hitting time  satisfies: $g^{(\mu+\lambda)}(x_0) \ge  g^{(1+1)}(x_0)/c$. Then we have $PS(\mu+\lambda \mid x_0) \le c.$ 
 If inequality (\ref{equPSUpperBound}) is strict,  then according to Theorem~\ref{theCompareLower}, $PS(\mu+\lambda \mid x_0) < c$. 
\end{IEEEproof} 
 
 The second  theorem establishes a sufficient condition     for estimating   the lower bound on population scalability.
\begin{theorem} 
\label{thePSLowerBound} Given a non-optimal  $x_0$, if  for all $t \ge 0$,  average drift 
\begin{align}
\label{equPSLowerBound}
\bar{\Delta}^{(\mu+\lambda)}_t     \ge c 
\end{align}
where $c >0$, then $PS(\mu+\lambda \mid x_0) \ge c.$ Furthermore if for at least one $t \ge 0$,  Inequality (\ref{equPSLowerBound}) is strict,  then $PS(\mu+\lambda \mid x_0)> c$. 
\end{theorem}

\begin{IEEEproof}
According to Theorem~\ref{theCompareUpper}, the expected hitting time  satisfies: $g^{(\mu+\lambda)}(x_0) \le  g^{(1+1)}(x_0)/c$. Then we have $PS(\mu+\lambda \mid x_0) \ge c.$ 
 If inequality (\ref{equPSLowerBound}) is strict,  then according to Theorem~\ref{theCompareUpper}, $PS(\mu+\lambda \mid x_0) >c$. 
\end{IEEEproof}

\subsection{Time-fitness landscape}
The analysis of population scalability is   established upon the  time-fitness landscape, a concept inroduced in \cite{he2015easiest}. It aims at describing the fitness landscape related to a general search space, a finite set which might be a set of  strings or  a set of graphs. 
\begin{definition}
Given a $(1+1)$ elitist EA for maximizing a function $f(x)$, its time-fitness landscape is the set of pairs $(g(x), f(x))$, where $g(x)$ is the expected hitting time of the EA starting at $x$. The  neighbour of $x$ includes  two points:  the point $y_1$ such that $g(y_1)$ is the closest to $g(x)$ from the direction $g(y) <g(x)$, and    the point $y_2$ such that $g(y_2)$ is the closest to $g(x)$ from the direction $g(y) >g(x)$. 
\end{definition}

The  time-fitness landscape  is competely different from traditional ones based on a distance. The formmer is related to a $(1+1)$ EA, but the latter usually not. Let's show the difference by unimodal functions.    A function is called unimodal if every non-optimal   point has a neighbour  with a strictly better fitness~\cite{rowe2012choice}. Traditionally  the definition of the neighbour  relies on a distance. For example, if  the search space is the set  of binary strings,   the neighbour   of  a point $x$ includes all points $y$ with Hamming distance 1 from $x$~\cite{rowe2012choice}. But such a definition is applicable to unimodal functions to a finite set because the distance is unknown. Thus  unimodal functions are defined  on the time-fitness landscape instead. 

\begin{figure}[ht]
\centering  
\begin{tikzpicture}
\begin{axis}[width=8cm,height=4.5cm,
xmax=7,ymax=14,
axis lines=left,
ticks=none, 
legend style={draw=none},
xlabel=$g^{(1+1)}(x)$, ylabel=$f(x)$,  
]  
\addplot+[nodes near coords,only marks,
point meta=explicit symbolic]
table[meta=label] {
x y label
0 10 $s_0$
1 9 $s_1$
2 6 $\,$
3 5 $\,$
4 4 $\,$
5 2 $s_{K-1}$
6 0 $s_K$
};
\end{axis}
\end{tikzpicture}
\caption{A unimodal time-fitness landscape. The $x$ axis is   the expected hitting time of the $(1+1)$ EA. The $y$ axis is the fitness function.}
\label{figUnimodal}
 \end{figure}
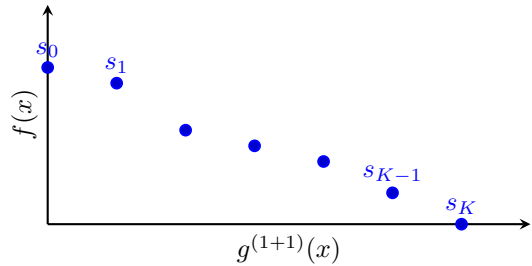

\begin{definition}
\label{defUnimodal}
Let $\mathcal{S}=\{ s_0, s_1, \cdots, s_K \}$ and the fitness function satisfy that
$
f(s_0)>f(s_{1}) >  \cdots > f(s_K).
$ 
Given a  $(1+1)$ elitist EA to maximise a  function $f$, $f$ is called    \emph{unimodal}  to the $(1+1)$ EA if   $g^{(1+1)}(s_1)< \cdots < g^{(1+1)}(s_{K})$.
\end{definition} 
A unimodal time-fitness landscape    is visualised in Fig.~\ref{figUnimodal}.
 
Unimodal functions in terms of the time-fitness landscape  are not equivalent to  unimodal functions in terms of the distance-based fitness landscape.  For example, a OneMax function below is unimodal under Hamming distance. 
\begin{equation}
f(x)=|x|, \quad x \in \{0,1\}^n,
\end{equation}  
where  $|x|$ denotes the number of 1-valued bits. Let $k$ represent the points $x$ such that $|x|=k$, and then the fitness function satisfies $f(n)>f(n-1)> \cdots>f(0)$. Consider a $(1+1)$ elitist EA as follows:
\begin{itemize}
\item \textbf{Mutation.}  
Mutation probabilities are: 
\begin{equation}
\left\{
\begin{array}{lll}
P_m(k,n)=\frac{1}{n \ln (2+k)}, \\
P_m(k,k) =1-P_m(k,k),
\end{array}\right.
   k=0,\cdots,n-1.
\end{equation}
 
\item  \textbf{Elitist Selection.}   Select the best individual  from $\Phi_t \cup \Psi_t$. 

\end{itemize} 

The expected hitting time satisfies 
$$
g^{(1+1)} (n-1) =n \ln (n+1) > \cdots > g^{(1+1)}(0)= n \ln 2.
$$
According to Definition~\ref{defUnimodal}, $f$ is  not unimodal to the  $(1+1)$ EA in terms of the time-fitness landscape.
 
A Fully-Deceptive function below is not unimodal under Hamming distance. 
\begin{equation}
f(x)=\left\{
\begin{array}{llll}
n, &\textrm{if } |x|=n, \\
n-1-|x|, &\textrm{if } |x| <n,
\end{array}
\right.
 x \in \{0,1\}^n,
\end{equation}  
The fitness function satisfies $f(n)>f(0)>\cdots>f(n-1)$. Still consider the same $(1+1)$ elitist EA  as above.
The expected hitting time still satisfies 
$$
g^{(1+1)} (0) =n \ln 2< \cdots < g^{(1+1)}(n-1)= n \ln (n+1).
$$
According to Definition~\ref{defUnimodal}, $f$ is    unimodal to the  $(1+1)$ EA in terms of the time-fitness landscape.

\section{Scenario 1: Using a population cannot reduce  hitting time}
\label{secAnalysis1}
\subsection{Case study 1: two-paths-I functions} 
\label{secTwoPathsI}
It is an intuitive  belief that  using a population may reduce  the number of generations to find an optimal solution.  The following case study shows this belief is not always true. 

Before the case study, a basic concept are revisited:    path~\cite{he1995convergence}\cite{he2015easiest}. Given a $(1+1)$   EA for maximizing   $f(x)$,  a path is a sequence of points $\{x_0 \to  x_1 \to \cdots \to  x_{k}\}$ such that $P(x_{i-1},x_i) >0$ for $i=1, \cdots, k$. The path is denoted by $\mathrm{Path}(x_0, x_1, \cdots, x_k)$. The case study is about  two-paths-I functions which are defined as below.

\begin{figure}[ht] 
\centering
\begin{tikzpicture}
\begin{axis}[width=8cm,height=4.5cm,
xmax=7,ymax=14,
axis lines=left,
ticks=none, 
legend style={draw=none},
xlabel=$g^{(1+1)}(x)$, ylabel=$f(x)$,  
] 
\addplot+[black, nodes near coords, 
point meta=explicit symbolic]
table[meta=label] {
x y label
0 10 $s_0$
1 4 $s_{1}$
2 2 $\,$ 
3 0 $s_{K}$
4 -2 $s_{K+L+1}$
}; 
\addplot+[black, nodes near coords, 
point meta=explicit symbolic,  dashed]
table[meta=label] {
x y label
0 10 $s_0$
1 8 $s_{K+1}$
2 7 $\,$
3 6.5 $\,$
5 5 $\,$
6 4 $s_{K+L}$
4 -2 $s_{K+L+1}$
};

\end{axis}
\end{tikzpicture}   
\caption{A two-paths-I time-fitness landscape. The $x$ axis represents   the expected hitting time    of the $(1+1)$ EA. The $y$ axis is the fitness function.   }
\label{figTwoPaths}
 \end{figure}
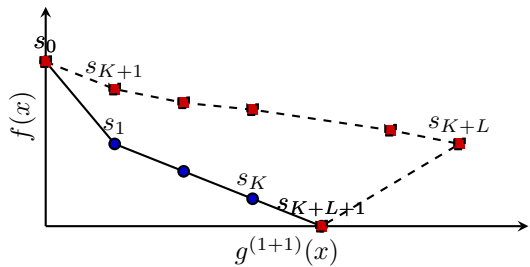

\begin{definition}  Let    $\mathcal{S}=\{ s_0, s_1,  \cdots,  s_{K+L} ,s_{K+L+1} \}$ where $L >K$. The fitness function satisfies that 
\begin{align}
&f(s_0)>f(s_{K+1}) >f(s_{K+2}) > \cdots > f(s_{K+L})\nonumber \\
>&f(s_{1}) >f(s_{2}) > \cdots > f(s_K) >f(s_{K+L+1}).
\end{align}  

Given a $(1+1)$ elitist EA to maximise a  function $f$, $f$ is called a \emph{two-paths-I function} to the $(1+1)$ EA if   there exist two paths  to  the optimum:  $\text{Path}_1(s_{K+L+1}, s_K,  \cdots, s_{1}, s_0)$ and $\text{Path}_2(s_{K+L+1}, s_{K+L}, \cdots, s_{K+1}, s_0)$ such that   
\begin{itemize}
\item for $k=1, \cdots, K$ and $k=K+2, \cdots, K+L$,  mutation probabilities    $ P_{\mathrm{m}}(s_k, s_{k-1})=1 $;
\item for $k=K+1$,  mutation probability   $ P_{\mathrm{m}}(s_k, s_0)=1$; 
\item  for $k=K+L+1$,  mutation probabilities  $ P(s_{k}, s_{K})=p$ and $ P_{\mathrm{m}}(s_{k}, s_{K+L})=1-p$ where $0<p<1$;
\item for any other $i,j$,  $ P_{\mathrm{m}}(s_i,s_j) =0$.
\end{itemize}   Fig.~\ref{figTwoPaths} visualises    a two-paths-I time-fitness landscape. 
\end{definition}

 Consider a $(1+\lambda)$ EA (where $\lambda \ge 2$) for maximizing a two-paths-I function function.  
\begin{itemize}
\item \textbf{Mutation.}  Mutation probabilities are identical to those in the $(1+1)$  EA. Generate $\lambda$ children.
 
\item  \textbf{Elitist Selection.}   Select the best individual  from $\Phi_t \cup \Psi_t$. 

\end{itemize}

The theorem below shows  that using a population will  increase the expected hitting time if the EA starts at $s_{K+L+1}$.

\begin{proposition}
\label{proTwoPathsI}
Given  the $(1+1)$  elitist EA and $(1+\lambda)$ EA (where $\lambda \ge 2$) for maximizing  a   two-paths-I  function,  let $\Phi^{(1+1)}_0=\Phi^{(1+\lambda)}_0= s_{K+L+1}$, then $PS(1+\lambda \mid s_{K+L+1}) <1$. The scalability threshold   is $2$.
\end{proposition}

\begin{IEEEproof}
For the $(1+\lambda)$ EA,  let its distance function  
$d^{(1+\lambda)} (x)=  g^{(1+1)}(x).
$
Since $\Phi^{(1+\lambda)}_0=s_{K+L+1}$ and $f(s_{K+L}) > f(s_K)$, there are two potential events.
\begin{enumerate}
\item The $(1+\lambda)$ EA moves  from $   s_{K+L+1}  $ to $ s_{K}  $. This event could happen if and only if all mutated children are $s_{K}$.  The probability  for the event happening is $p^{\lambda}$.

\item The $(1+\lambda)$ EA moves  from $ s_{K+L+1}$ to $s_{K+L}$. This event could happen if and only if at least one mutated child is $s_{K+L}$.   The probability  for the event happening is $1-p^{\lambda}$.
\end{enumerate}

We calculate the pointwise drift at $s_{K+L+1}$ as follows:
\begin{align} 
&\Delta^{(1+\lambda)}( s_{K+L+1})\nonumber \\
=& (1-p^{\lambda} )( g^{(1+1)} (s_{K+L+1}) -g^{(1+1)}(s_{K+L})) \nonumber\\
&+ p^{\lambda} (g^{(1+1)} (s_{K+L+1})- g^{(1+1)} (s_{K} ))\nonumber\\
= & (1-p^{\lambda} )[1+pK+(1-p)L -L] \nonumber\\
&+p^{\lambda} [1+pK+(1-p)L-K] \nonumber \\
=&1+(p-p^{\lambda})(K-L)<1.  \\
& (\text{since }L>K,  p \in (0,1) \text{ and }\lambda \ge 2  )  \nonumber
\end{align}
We calculate the pointwise drift at any other non-optimal $s \in \{s_1, \cdots, s_{K+L}\}$ . The pointwise drift equals to  $1 $.
 
Since $\Phi_0^{(1+\lambda)}=s_{K+L+1}$,  average drift $\bar{\Delta}^{(1+\lambda)}_0  <1$. For any $t\ge 1$, $\Phi^{(1+\lambda)}_t$ has left the point $s_{K+L+1}$, then   average drift
$ 
\bar{\Delta}^{(1+\lambda)}_t  = 1 
$. According to Theorem~\ref{thePSUpperBound},   we get that  $PS(1+\lambda \mid s_{K+L+1})<1$.
\end{IEEEproof}
It is easy to understand the proposition. There are two paths towards the optimum: short and long.   If using a population,  the long path is more likely to be chosen than the short path. Then the expected hitting time is increased.  

Proposition~\ref{proTwoPathsI} implies that even if let the population size $\lambda \to +\infty$, the expected hitting time of $(1+\lambda)$ EA is still larger than that of the $(1+1)$ EA on two-paths-I functions.

\begin{example}
\label{exaTwoPathsI}
Consider an instance of two-paths-I functions. Let $x \in \{0,1\}^n$ and $|x|$ denote its number of 1-valued bits. 
\begin{equation}
\label{equTwoPathsI}
 f(x)=\left\{
\begin{array}{lll}
n, & \text{if } |x|=0 \text{ or } |x| = n,\\
-|x|, & \text{if } |x| \le \theta \text{ where }\theta=2,\\
|x|, & \text{otherwise}.
\end{array}
\right.  
\end{equation}
There are two optima: $0\cdots0$ and $1\cdots 1$. Let $x_0$ be a string with  $|x_0|=2$. It takes the minimum value of the fitness. A $(1+\lambda)$ EA  uses adaptive mutation and elitist selection for solving the above problem.
\begin{itemize} 
\item  \textbf{Adaptive mutation.}   
\begin{enumerate}
\item  When $t =0$,   flip one of  1-valued bits with the probability 0.5, otherwise flip one of 0-valued bits. In this way, generate $\lambda$ children as $\Psi_0$. 

\item When $t \ge 1$, if $f(\Phi_t) >f(\Phi_{t-1})$ and $\Phi_t$ is generated by flipping a 0-valued bit (or 1-valued) in $\Phi_{t-1}$,   flip   a 0-valued bit (or 1-value bit). Then generate $\lambda$ children as $\Psi_t$. 

\item  When $t \ge 1$, if $f(\Phi_t)= f(\Phi_{t-1})$,  either  flip one of  1-valued bits with probability 0.5 or  flip one of 0-valued bits.  Then generate $\lambda$ children as $\Psi_t$. 
\end{enumerate}  
\item  \textbf{Elitist Selection.}   Select the best individual  from $\Phi_t \cup \Psi_t$ as $\Phi_{t+1}$.   
\end{itemize}

   Table~\ref{tabTwoPathsI} shows that using a  population   increases the expected hitting time.

\begin{table}[htb]
\caption{Experimental results for Example~\ref{exaTwoPathsI} averaged over 1000 runs. $n=1000$. The   EA   starts at $x_0$ with  $|x_0|=2$.}
\label{tabTwoPathsI}
\centering
\begin{tabular}{c|c|c|c|c|c }  
    \toprule
    population size   & 1 &  3 & 5   &7 &9   \\
    \midrule
    hitting time  & 491& 872& 961& 990& 997 \\ 
    \midrule
    running time   & 491& 2616& 4805& 6930& 8973\\ 
    \bottomrule
\end{tabular}
\end{table}

\end{example}

Notice that the function in the  example  is a unimodal function under Hamming distance but not unimodal in terms of the time-fitness landscape (Fig.~\ref{figTwoPaths}).

\subsection{Case study 2: unimodal functions}
Here is another case study to show that   using a population can not reduce  the number of generations to find an optimal solution. 
The case study discusses a $(\mu+1)$ EA (where $\mu \ge 1$) with elitist selection for maximizing any unimodal function.  
\begin{itemize}
\item    \textbf{Global Mutation.}  Choose  one individual from $\Phi_t$ at random and generate a child by mutation. Mutation probability $ P_{\mathrm{m}}(s_i,s_j)>0$ for any $i$ and $j$.

\item  \textbf{Elitist Selection.} If the child's fitness is better than  that of one or more parents, then the child will replace one of these parents at random. 
\end{itemize}

The proposition below asserts that using a population  increases the expected hitting time of the $(\mu+1)$ EA. 
\begin{proposition}
\label{proUnimodal1}
Given  the   $(1+1)$   EA and $(\mu+1)$ EA (where $\mu \ge 2$) for maximizing  a unimodal  function,
for any $x \in \{ s_2, \cdots, s_K\}$, 
let $\Phi^{(1+1)}_0=x$   and $\Phi^{(\mu+1)}_0=(x,\cdots, x)$,  then $PS(\mu+1 \mid x)<1$.  The scalability threshold  is $2$.
\end{proposition}

\begin{IEEEproof}
For the $(1+1)$  EA,  choose $
 g^{(1+1)}  (x)$  to be  its distance function.
According to Theorem~\ref{theFundamentalMatrix}, the pointwise drift satisfies 
$ \Delta^{(1+1)} (s_k)  
=  1 
$ for any $s_k \in \{ s_1, \cdots, s_K\}$.
Since selection is elitist and the function is unimodal, we have  
\begin{align}
\label{equDriftOne2}
 \Delta^{(1+1)} (s_k)  = \scriptstyle  \sum^{k-1}_{l=0}  P_{\mathrm{m}} (s_k,s_l)\left(  g^{(1+1)}  (s_k) -  g^{(1+1)}  (s_l)   \right)  
=1.
\end{align}

For the  $(\mu+1)$ EA, define its distance function  
$
d^{(\mu+1)}( X) =  \min \{  g^{(1+1)}(x):  x \in X \}.
$

Let $\Phi^{(\mu+1)}_t=  X=(s_{1(X)}, \cdots, s_{\mu(X)})$ whose best individual is $s_k$, and  $ \Psi^{(\mu+1)}_t= s_l$.   Since the $(\mu+1)$ EA adopts  elitist selection and the fitness function is unimodal, the pointwise drift satisfies
\begin{align}
 \Delta^{(\mu+1)} (X)   
=& \scriptstyle\sum^{k-1}_{l=0}  P_{\mathrm{m}}(X,s_l) \left (   g^{(1+1)}(s_k)  -  g^{(1+1)} (s_l) \right).\label{equDriftUni2}
\end{align}

The probability of mutating  $s_{i(X)}$ to $s_l$  is
$ 
 P_{\mathrm{m}}(s_{i(X)},s_l).
$ 
 Since each parent is chosen for mutation at random,   we have
$$
 P_{\mathrm{m}}(X,s_l)  =\frac{1}{\mu} \sum^{\mu}_{i=1}  P_{\mathrm{m}}(s_{i(X)},s_l).
$$
Then we get
\begin{align*}
 &  P_{\mathrm{m}}(X,s_l) \left (   g^{(1+1)}(s_k)  -  g^{(1+1)} (s_l) \right)  \nonumber \\
=& \textstyle \frac{1}{\mu} \sum^{\mu}_{i=1}  P_{\mathrm{m}}(s_{i(X)},s_l) (    g^{(1+1)}(s_k) -     g^{(1+1)}(s_l)  ) .
\end{align*}

Since $l <k \le i(X)$, we have 
$$    g^{(1+1)}(s_l) < g^{(1+1)}(s_{k}) \le  g^{(1+1)}(s_{i(X)}), 
$$
 and then
\begin{align*}
 & P_{\mathrm{m}}(X,s_l) \left (   g^{(1+1)}(s_k)  -  g^{(1+1)} (s_l) \right)  \nonumber \\ 
\le & \textstyle \frac{1}{\mu} \sum^{\mu}_{i=1}  P_{\mathrm{m}}(s_{i(X)},s_l) (    g^{(1+1)}(s_{i(X)}) -     g^{(1+1)}(s_l)  )
\end{align*}
 
The pointwise  drift satisfies
\begin{align}
 & \Delta^{(\mu+1)} (X)\nonumber \\ 
=& \scriptstyle \frac{1}{\mu} \sum^{\mu}_{i=1} \sum^{k-1}_{l=0}   P_{\mathrm{m}}(s_{i(X)},s_l) (    g^{(1+1)}(s_k) -     g^{(1+1)}(s_l)  ) \nonumber \\
\le &\scriptstyle \frac{1}{\mu} \sum^{\mu}_{i=1}   \sum^{k-1}_{l=0} P_{\mathrm{m}}(s_{i(X)},s_l) (    g^{(1+1)}(s_{i(X)}) -     g^{(1+1)}(s_l)  ).
\end{align} 
 
 \begin{enumerate}
 \item 
\textbf{Case 1:}   for all individuals in $X$,  $s_{i(X)} = s_k$. 

In this case, according to (\ref{equDriftOne2}), we have 
\begin{align*}
\textstyle\sum^{k-1}_{l=0} P_{\mathrm{m}}(s_k,s_l) (    g^{(1+1)}(s_k) -     g^{(1+1)}(s_l)  )=1.
\end{align*}
Then the pointwise  drift satisfies
\begin{align} \Delta^{(\mu+1)} (X)  
= \frac{1}{\mu} \sum^{\mu}_{i=1}  1=1 .
\end{align}

\item \textbf{Case 2:}  for  at least one of individuals in $X$, $s_{i(X)} \neq s_k$. 

 Since $s_k$ is the best individual in $X$, the indexes satisfy $i(X) >k$. 
Thanks to global mutation, we have $P_{\mathrm{m}}(s_{i(X)}, s_l)>0$ for any  $i(X)$, and then  
\begin{align}
& \scriptstyle\sum^{k-1}_{l=0}  P_{\mathrm{m}}(s_{i(X)}, s_l) (   g^{(1+1)}(s_{i(X)}) -    g^{(1+1)}(s_l) )  \nonumber  \\ 
<& \scriptstyle\sum^{i(X)-1}_{l=0}  P_{\mathrm{m}}(s_{i(X)}, s_l) (   g^{(1+1)}(s_{i(X)}) -    g^{(1+1)}(s_l) )  \nonumber  \\
=&\Delta^{(1+1)}(s_{i(X)})=1.
\label{equDriftInequ}
\end{align}  
The last equality ``$=1$'' is based on (\ref{equDriftOne2}).
Then the pointwise  drift satisfies
\begin{align} 
\Delta^{(\mu+1)} (X)  
< \frac{1}{\mu} \sum^{\mu}_{i=1}  1=1.
\end{align} 
\end{enumerate}

Since $\Phi_0=(s_k, \cdots, s_k)$ where $k \ge 2$, the  average drift $\bar{\Delta}^{(\mu+1)}_0 =1$. When $t=1$,  the probability  of $\Phi_1 \in \mathcal{P}_{\non}$ including two different non-optimal points (say $s_k$ and $s_{k-1}$) is always greater than $0$ due to global mutation. Thus  the average drift $\bar{\Delta}^{(\mu+1)}_1  <1$. When $t \ge 2$, the average drift $\bar{\Delta}^{(\mu+1)}_t   \le 1$.
 According to Theorem~\ref{thePSUpperBound},   we get  $PS(\mu+1\mid s_k)<1$.
\end{IEEEproof}

Here is an explanation of  this proposition. For unimodal functions, the higher the fitness of a point is, the closer to the optimal set the point is (Fig.~\ref{figUnimodal}). Given a population in the $(\mu+1)$ EA, a good strategy is to mutate the best individual in the population.  Unfortunately, a population may include one or more individuals which are worse than the best. If EA chooses a worse individual to mutate, it will increase he expected hitting time.

Proposition~\ref{proUnimodal1} shows that even if let the population size $\mu \to +\infty$, the expected hitting time of the $(\mu+1)$ EA is still larger than that of the $(1+1)$ EA on unimodal functions. 

\begin{example}
\label{exaTwoMax0}
Consider an instance of unimodal functions. Let $x \in \{0,1\}^n$ and $|x|$ denote the number of 1-valued bits.  
\begin{equation}
\label{equTwoMax0}
   f(x)=  \max \{ |x|, n-|x| \}.
\end{equation}
There are two optima: $0\cdots0\vee1\cdots 1$. A $(\mu+1)$ elitist EA  is used for solving the maximization problem.
\begin{itemize}
\item    \textbf{Bitwise Mutation.} Choose one individual from $\mu$ parents at random. Flip each bit with a probability $1/n$. 

\item  \textbf{Elitist Selection.} If the child's fitness is better than  that of one or more parents, then the child will replace one of these parents at random.   
\end{itemize}
\end{example} 

Table~\ref{tabTwoMax0}  shows that using a population  increases the expected  hitting time.
\begin{table}[htb]
\caption{Experimental results for Example~\ref{exaTwoMax0} averaged over 1000 runs. $n=200$. The EA starts at $x_0$ with  $|x_0| =100$.}
\label{tabTwoMax0}
\centering
\begin{tabular}{c|c|c|c|c|c }  
    \toprule
    population size   & 1 &  3 & 5   &7 &9   \\
    \midrule
    hitting time   & 2523& 2650& 3014& 3477& 3855 \\ 
    \midrule
    running time    & 2523& 2650& 3014& 3477& 3855 \\ 
    \bottomrule
\end{tabular}
\end{table}

\section{Scenario 2: Using a population can reduce  hitting time but not   running time }
\label{secAnalysis2}  
\subsection{Case study 3: unimodal functions }
Let's reinvestigate the intuitive  belief that using a population   can reduce  the expected hitting time    of  elitist EAs for maximizing   unimodal functions. Although this belief is not  true for the $(\mu+1)$ EA, it is still  true for the $(\lambda+\lambda)$ EA with global mutation and elitist selection.
     
Consider a $(\lambda+\lambda)$ EA $(\lambda\ge 1)$ using elitist selection and  global mutation. 
\begin{itemize}
\item    \textbf{Global Mutation.} For any   $i,j$, mutation probability $ P_{\mathrm{m}}(s_i,s_j)>0$.  Each individual in $\Phi_t$ generates a child. 

\item  \textbf{Elitist Selection.} Probabilistically select $\lambda$ individuals   from $\Phi_t \cup \Psi_t$, while the best individual is always selected.
\end{itemize}

First we prove an inequality which will be used later.
\begin{lemma} 
\label{lemInq}
Given $a_i>0, b_i>0, c_i>0$ where $i=0,1,\cdots, k$ such that
 $\sum^j_{i=0} a_i>\sum^j_{i=0} b_i, $ $   j=0, \cdots, k$ and $c_0>c_1 > \cdots > c_k$,
it holds $\sum^k_{i=0} a_i c_i > \sum^k_{i=0} b_i c_i$.
\end{lemma}

\begin{IEEEproof}
  From the conditions $a_0 +\cdots +a_j > b_0+\cdots+b_j$ and  $c_0 > c_1 > \cdots >c_k$, we have
\begin{align*}\begin{array}{llll}
&\sum^k_{i=0} (a_i -b_i)c_i \nonumber\\
=&(a_0-b_0) c_0 + \sum^k_{i=1} (a_i -b_i)c_i  \nonumber\\
>& (a_0-b_0) c_1 + \sum^k_{i=1} (a_i -b_i)c_i  \nonumber\\
=&  (a_0-b_0+a_1-b_0) c_1 + \sum^k_{i=2} (a_i -b_i)c_i \nonumber\\
>&(a_0-b_0+a_1-b_1) c_2+ \sum^k_{i=2} (a_i -b_i)c_i \nonumber\\
=&  (a_0-b_0+a_1-b_0+a_2-b_2) c_2 + \sum^k_{i=3} (a_i -b_i)c_i.
\end{array}
\end{align*}
By induction, we can prove that
\begin{align*}
\begin{array}{lll}
&\sum^k_{i=0} (a_i -b_i)c_i  \nonumber\\
>&(a_0-b_0+a_1-b_1 +\cdots +a_k - b_k) c_k >0.
\end{array}
\end{align*}
This gives the desired result.
\end{IEEEproof}

\begin{proposition}
\label{proUnimodal2} 
Given  the $(1+1)$  elitist EA and $(\lambda+\lambda)$ EA (where $\lambda \ge 2$) using global mutation and elitist selection for maximizing  any unimodal function,  for any $x \in \mathcal{P}_{\non}$, 
let $\Phi^{(1+1)}_0=x$   and $\Phi^{(\lambda+\lambda)}_0=(x,\cdots, x)$, then $PS(\lambda+\lambda \mid x)>1$.  
\end{proposition}
 
\begin{IEEEproof}
For the $(1+1)$  EA,  choose $
 g^{(1+1)}  (x)
$  to be  its distance function.
According to Theorem~\ref{theFundamentalMatrix}, the pointwise drift satisfies 
$ \Delta^{(1+1)} (s_k)  
=  1
$ for any $s_k \in \mathcal{P}_{\non}$.
Since selection is elitist and the function is unimodal, we have
\begin{align} 
\Delta^{(1+1)} (x)  = \scriptstyle \sum^{k-1}_{l=0}  P_{\mathrm{m}}(s_k,s_l)  (  g^{(1+1)}  (s_k) -  g^{(1+1)}  (s_l)    )= 1. 
\end{align}

For a  $(\lambda+\lambda)$ EA, define its distance function 
$ 
d^{(\lambda+\lambda)}( X) =  \min \{ g^{(1+1)}(x): x \in X \}.
$ 

Let $\mathcal{P}_k$ denote the set of populations whose best individual is $s_k$ (where $k=1 \cdots, K$). 
Assume that $ \Phi^{(\lambda+\lambda)}_t=  X   \in \mathcal{P}_{k}$ (where $k >0$), $ \Psi^{(\lambda+\lambda)}_t=Y $ and $ \Phi^{(\lambda+\lambda)}_{t+1}=Z \in \mathcal{P}_m$ (where $m< k$). 

Given any $m <  k$, population  $\Phi_{t+1}$ enters in the union  $\mathcal{P}_0 \cup  \mathcal{P}_1 \cup \cdots \cup \mathcal{P}_m$  if and only if one or more individuals in population $\Phi_t$ is mutated into an individual in the set $\{s_0, \cdots, s_{m}\}$. Thanks to global mutation and population size $\lambda \ge 2$, this probability is strictly larger than that of one individual ($s_k$) being mutated into   the set $\{s_0, \cdots, s_{m}\}$, that is
\begin{align}
& \sum^m_{l=0} \sum_{Z \in \mathcal{P}_l } P(X, Z) 
>  \sum^m_{l=0}  P(s_k, s_l).\nonumber
\end{align}
Since the fitness function is unimodal, we have 
\begin{align*}
&g^{(1+1)}(s_k)-g^{(1+1)}(s_0) 
> g^{(1+1)}(s_k)-g^{(1+1)}(s_1)  \\
&> \cdots >  g^{(1+1)}(s_k)-g^{(1+1)}(s_{k-1}). 
\end{align*}

Using Lemma~\ref{lemInq} (let $a_l =\sum_{Z \in \mathcal{P}_l } P(X, Z) $, $b_l =P(s_k, s_l)$ and $c_l= g^{(1+1)}(s_k)-g^{(1+1)}(s_l)$), we get
\begin{align*}  
 \Delta^{(\lambda+\lambda)} (X) 
=& \sum^{k-1}_{l=0}\sum_{Z \in \mathcal{P}_l} P(X, Z)  (g^{(1+1)}(s_k)-g^{(1+1)}(s_l))\nonumber\\
> &  \sum^{k-1}_{l=0} P(s_k, s_l)  (g^{(1+1)}(s_k)-g^{(1+1)}(s_l))\nonumber \\
 =&
\Delta^{(1+1)} (s_k)=1 .
\end{align*}
Then we have
$
\Delta^{(\lambda+\lambda)} (X)  >  1.
$ Since $\Phi_0 \in \mathcal{P}_{\non}$, we have  for any $t \ge 0$, the average drift $\bar{\Delta}^{(\lambda+\lambda)}_t >1.$
Applying Theorem \ref{thePSLowerBound}, we get  $PS(\lambda+\lambda)>1$.  
\end{IEEEproof}

Proposition~\ref{proUnimodal3} states that using a  population   can reduce the expected hitting time of the   $(\lambda+\lambda)$ EA on unimodal  functions.   
Here is an explanation.  For unimodal functions, the higher the fitness of a point is, the closer  to the optimal set the point is. The probability of the $(\lambda+\lambda)$ EA (where $\lambda\ge 2$) to generate a better individual is strictly larger than that of the $(1+1)$ EA. Thus the expected hitting time is shortened.

\subsection{Case Study 4: unimodal functions}
It is an intuitive  belief that using a population   can not reduce the expected  running time  of   elitist EAs for maximizing unimodal functions.
 The proposition below asserts this is true for unimodal functions in terms of the time-fitness landscape. 
 
Consider a   $(\mu+\lambda)$ elitist EA.
\begin{itemize}
\item    \textbf{Mutation.}  Select $\lambda$ individuals in $\Phi_t$ and mutate them. Then generate a children population consisting of $\lambda$  individuals;

\item  \textbf{Elitist Selection.} First select one individual with the highest fitness in $\Phi_t \cup \Psi_t$; and then probabilistically select $\mu-1$ individuals   from $\Phi_t \cup \Psi_t$.
\end{itemize}

\begin{proposition}
\label{proUnimodal3}
Given  the   $(1+1)$  elitist EA and $(\mu+\lambda)$ EA (where  $\mu \ge 2$ or $ \lambda \ge 2$) for maximizing  a  unimodal  function,  for any $x \in \mathcal{P}_{\non}$, 
let $\Phi^{(1+1)}_0=x$   and $\Phi^{(\mu+\lambda)}_0=(x,\cdots, x)$,  then $PS(\mu+\lambda \mid x)<\lambda$.  
\end{proposition}

\begin{IEEEproof}
It is sufficient to consider the case of $(\mu+\lambda)$ with $\lambda >1$. The analysis of $(\mu+1)$ EA is almost the same as that of  Proposition~\ref{proUnimodal1}, except two places: (1) without the global mutation condition, inequality (\ref{equDriftInequ})  is changed from $<$  to $\le $; (2) the  conclusion is changed from  $PS(\mu+1) < 1$ to $PS(\mu+1) \le 1$.

For the $(1+1)$  EA,  choose $
 g^{(1+1)}  (x)$  to be  its distance function.
According to Theorem~\ref{theFundamentalMatrix}, for any $s_k \in \{ s_1, \cdots, s_K \}$,  the pointwise drift satisfies 
$  \Delta^{(1+1)} (s_k) =   1. 
$  
Since selection is elitist and the function is unimodal, we have  
\begin{align}
\label{equDriftOne} 
\Delta^{(1+1)} (s_k)  = \scriptstyle \sum^{k-1}_{l=0}  P_{\mathrm{m}}(s_k,s_l)  (  g^{(1+1)}  (s_k) -  g^{(1+1)}  (s_l)    )= 1. 
\end{align}

For the  $(\mu+\lambda)$ EA where $\lambda \ge 2$, let its distance function  
$
d^{(\mu+\lambda)}( X) =  \min \{  g^{(1+1)}(x):  x \in X \}.
$
 
Let $\mathcal{P}_k$ denote the set of populations whose best individual is $s_k$ (where $k=0,1,\cdots, K$). 
Assume that $ \Phi^{(\mu+\lambda)}_t=  X   \in \mathcal{P}_{k}$ where $k \ge 1$, $ \Psi^{(\mu+\lambda)}_t=Y $ and $ \Phi^{(\mu+\lambda)}_{t+1}=Z  \in \mathcal{P}_{l}$. 

 Since the $(\mu+\lambda)$ EA adopts  elitist selection and the fitness function is unimodal,  the pointwise drift satisfies 
\begin{align}
&  \Delta^{(\mu+\lambda)}(X)  \nonumber \\
=&\scriptstyle \sum^{k-1}_{l=0} \sum_{Z \in \mathcal{P}_l} P(X,Z) \left (   g^{(1+1)}(s_k)  -  g^{(1+1)} (s_l) \right)\nonumber\\
=&\scriptstyle\sum^{k-1}_{l=0} \sum_{Y \in \mathcal{P}_l} P_{\mathrm{m}}(X,Y) \left (   g^{(1+1)}(s_k)  -  g^{(1+1)} (s_l) \right)  \label{equPointWise1}.
\end{align}

Denote children population $Y$ by $(s_{1(Y)}, \cdots, s_{\lambda(Y)})$. Let $(s_{1(X)}, \cdots, s_{\lambda(X)})$ be the parents from which  $Y$ are mutated.  Population  $Y \in \mathcal{P}_l$ (where $l <k$) only if one or more  parents is muted into $s_l$.  The probability of mutating   $s_{i(X)}$ to $s_l$ is
$ 
 P_{\mathrm{m}}(s_{i(X)},s_l).
$ 
 Since each individual is mutated independently,    the probability of one or more  parents is muted into $s_l$  is  not more than the sum of each parents is mutated into $s_l$.  
Then we have
\begin{align}
 &  \textstyle \sum_{Y \in \mathcal{P}_l} P_{\mathrm{m}}(X,Y) \left (   g^{(1+1)}(s_k)  -  g^{(1+1)} (s_l) \right)  \nonumber \\
\le & \textstyle  \sum^{\lambda}_{i=1}  P_{\mathrm{m}}(s_{i(X)},s_l) (    g^{(1+1)}(s_k) -     g^{(1+1)}(s_l)  ) .
\end{align}
The above inequality is strict  if $X=(s_k, \cdots, s_k)$. 
 
 Since $l <k \le i(X)$, we have 
$$    g^{(1+1)}(s_l) < g^{(1+1)}(s_{k}) \le  g^{(1+1)}(s_{i(X)}), 
$$
and then 
\begin{align*}
 & \textstyle \sum_{Y \in \mathcal{P}_l} P_{\mathrm{m}}(X,Y) \left (   g^{(1+1)}(s_k)  -  g^{(1+1)} (s_l) \right)  \nonumber \\
\le  & \textstyle   \sum^{\lambda}_{i=1}  P_{\mathrm{m}}(s_{i(X)},s_l) (    g^{(1+1)}(s_{i(X)}) -     g^{(1+1)}(s_l)  ) .
\end{align*} 

Inserting the above inequality into (\ref{equPointWise1}), we get
\begin{align}
& \Delta^{(\mu+\lambda)}(X)  \nonumber \\ 
\le &\scriptstyle\sum^{k-1}_{l=0} \sum^{\lambda}_{i=1}  P_{\mathrm{m}}(s_{i(X)},s_l) (    g^{(1+1)}(s_{i(X)}) -     g^{(1+1)}(s_l)  )  \nonumber\\
=&\scriptstyle \sum^{\lambda}_{i=1} \sum^{k-1}_{l=0} P_{\mathrm{m}}(s_{i(X)},s_l) (    g^{(1+1)}(s_{i(X)}) -     g^{(1+1)}(s_l)  )  \label{equDriftUnimodal}.
\end{align}

Since $s_k$ is the best individual in $X$, we have for $i=1, \cdots, \lambda$, the indexes satisfy  $i(X) \ge k$. Then
\begin{align*}
&\textstyle \sum^{k-1}_{l=0} P_{\mathrm{m}}(s_{i(X)},s_l) (    g^{(1+1)}(s_{i(X)}) -     g^{(1+1)}(s_l)  ) \\
\le & \textstyle\sum^{i(X)-1}_{l=0} P_{\mathrm{m}}(s_{i(X)},s_l) (    g^{(1+1)}(s_{i(X)}) -     g^{(1+1)}(s_l)  )  
=  1.  
\end{align*}
 
The drift satisfies
\begin{align*}
 \Delta^{(\mu+\lambda)}  ( X)  
\le  \sum^{\lambda}_{i=1} 1= \lambda.
\end{align*}
The above inequality is strict  if $X=(s_k, \cdots, s_k)$. 

Since $\Phi_0=(s_k, \cdots, s_k)$ for some $k \ge 1$,  the average drift $\bar{\Delta}^{(\mu+\lambda)} _0 <1$. When $t \ge 1$,   the average drift $\bar{\Delta}^{(\mu+\lambda)} _t \le 1$. Applying Theorem~\ref{thePSUpperBound}, we obtain 
 $PS(\mu+\lambda)<\lambda$.  
\end{IEEEproof}

Proposition~\ref{proUnimodal3} states that using a  population    cannot reduce the expected running time of the   $(\mu+\lambda)$ EA on unimodal  functions.   
The explanation is simple.  For unimodal functions, the higher the fitness of a point is, the closer  to the optimal set the point is. The probability of the $(\mu+\lambda)$ EA to generate a better individual is not more than $\lambda$ times that of the $(1+1)$ EA. Thus the expected hitting time cannot be shortened by $1/\lambda$.

\begin{example}
\label{exaTwoMax}
Consider the instance of  unidmodal functions in Example~\ref{exaTwoMax0} again.  
\begin{equation}
\label{equTwoMax}
   f(x)=  \max \{ |x|, n-|x| \}.
\end{equation}   A $(1+\lambda)$  EA  (where $\lambda\ge 1$) with  elitist selection and bitwise mutation is used for the maximizing the function.   
\begin{itemize}
\item    \textbf{Bitwise Mutation.} Flip each bit with a probability $1/n$. Then generates $\lambda$ children.   

\item  \textbf{Elitist Selection.} Select the best individual from $\Phi_t \cup \Psi_t$.
\end{itemize}

Table~\ref{tabTwoMax}  shows that using a population  reduces the expected  hitting time, but increases the expected   running time. 

\begin{table}[ht]
\caption{Experimental results for Example~\ref{exaTwoMax} averaged over 1000 runs. $n=200$. The EA starts at $x_0$ with  $|x_0| =100$.}\label{tabTwoMax}
\centering
\begin{tabular}{c|c|c|c|c|c }  
    \toprule
    population size   & 1 &  3 & 5   &7 &9   \\
    \midrule
    hitting time    & 2536& 864& 529& 395& 315 \\ 
    \midrule
    running time    & 2536& 2592& 2645& 2765& 2835  \\ 
    \bottomrule
\end{tabular}
\end{table}

\end{example}

Let's apply Proposition~\ref{proUnimodal3} to a special instance: the $(1+\lambda)$ EA (using bitwise mutation and elitist selection) for maximizing the OneMax function.  According to Proposition~\ref{proUnimodal3}, using a population will increase the running time. The population threshold is $2$. This  conclusion is different from that in \cite{jansen2005choice}. The   result  in \cite{jansen2005choice} asserts that the expected running time of the $(1+\lambda)$ EA  is the same order of that of the $(1+1)$ EA by a constant factor   when $\lambda$ is smaller than the cut-off point. The constant factor could be any constant such as  $1/1000$ or $1000$.  But the two results are not contrary. Our result indicates that the factor is less than $1$ when $\lambda\ge 2$.  

\section{Scenario 3:Using a population can reduce   running time}
\label{secAnalysis3}
\subsection{Case study 5: two-paths-II functions}
\label{secTwoPathsII}
 
In the previous section, it has been proven that using a population   cannot reduce the expected running time  of an EA for maximizing any unimodal function  in terms of the time-fitness landscape. But this intuitive belief  is not true in terms of the distance-based fitness landscape. The following case study of two-paths-II functions and its instance demonstrate this.
 
\begin{figure}[ht] 
\centering

\begin{tikzpicture}
\begin{axis}[width=8cm,height=5cm,
xmax=7,ymax=14,
axis lines=left,
ticks=none, 
legend style={draw=none},
xlabel=$g^{(1+1)}(x)$, ylabel=$f(x)$,  
]  
\addplot+[black,nodes near coords, 
point meta=explicit symbolic ]
table[meta=label] {
x y label
0 10 $s_0$
1 4 $s_1$
2 3 $\,$
3 2.5 $\,$
5 1 $s_{K-1}$
6 0 $s_{K}$
4 -2 $s_{K+L+1}$
};
\addplot+[black,nodes near coords, 
point meta=explicit symbolic,  dashed]
table[meta=label] {
x y label
0 10 $s_0$
1 9 $s_{K+1}$
2 7 $\,$ 
3 6 $s_{K+L}$
4 -2 $s_{K+L+1}$
};
\end{axis}
\end{tikzpicture}
 
\caption{A two-paths-II time-fitness landscape. The $x$ axis represents   the expected hitting time    of the $(1+1)$ EA. The $y$ axis is the fitness function.   }
\label{figTwoPathsII}
 \end{figure}
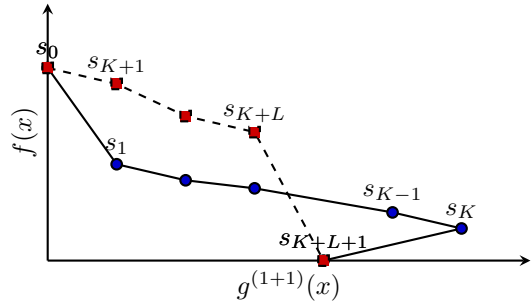

\begin{definition}  Let  $\mathcal{S}=\{ s_0, s_1,  \cdots,  s_{K+L} ,s_{K+L+1} \}$ where $L <K$ and the fitness function  $f(x)$ satisfy that
\begin{align}
&f(s_0)>f(s_{K+1}) >f(s_{K+2}) > \cdots > f(s_{K+L})\nonumber \\
>&f(s_{1}) >f(s_{2}) > \cdots > f(s_K) >f(s_{K+L+1}).
\end{align}  
Given a $(1+1)$ elitist EA to maximise  $f$, $f$ is called a two-paths-II function to the $(1+1)$ EA if 
there exist two paths  to  the optimum:  $\text{Path}_1(s_{K+L+1}, s_K,  \cdots, s_{1}, s_0)$ and $\text{Path}_2(s_{K+L+1}, s_{K+L}, \cdots, s_{K+1}, s_0)$ such that   
\begin{itemize}
\item for $k=1, \cdots, K$ and $k=K+2, \cdots, K+L$,  mutation probabilities    $ P_{\mathrm{m}}(s_k, s_{k-1})=1 $;
\item for $k=K+1$,  mutation probability   $ P_{\mathrm{m}}(s_k, s_0)=1$;
\item  for $k=K+L+1$,  mutation probabilities  $ P(s_{k}, s_{K})=p$ and $ P_{\mathrm{m}}(s_{k}, s_{K+L})=1-p$ where $0<p<1$;
\item for any other  $i,j$,  $ P_{\mathrm{m}}(s_i,s_j) =0$.
\end{itemize}    Fig.~\ref{figTwoPathsII} visualises   the a two-paths-II time-fitness landscape.
\end{definition}

 Consider a $(1+\lambda)$ EA where $\lambda \ge 2$.  
\begin{itemize}
\item \textbf{Mutation.} Mutation probabilities are identical to those in the $(1+1)$  EA in the above definition.

\item  \textbf{Elitist Selection.}   Select the best individual  from $\Phi_t \cup \Psi_t$. 

\end{itemize}

  Under certain condition, using a population may reduce the expected hitting time.

\begin{proposition}
\label{proTwoPathsII}
Given  the  $(1+1)$   EA and $(1+\lambda)$ EA (where $\lambda\ge 2$) for maximizing  a two-paths-II function, let $\Phi^{(1+1)}_0=s_{K+L+1}$ and $\Phi^{(1+\lambda)}_0=  s_{K+L+1}$, if the  population size  satisfies $\lambda< \lambda^*$ where $\lambda^*$ is given by
\begin{align}
\label{equConditionTwoPathsII}
 \lambda^*= 1+(p-\lambda p^{\lambda}) \frac{K-L}{L+1},
\end{align} then $PS(1+\lambda \mid s_{K+L+1}) >\lambda$. The scalability threshold is not less than $\lambda^*$.
\end{proposition}

\begin{IEEEproof}
For the $(1+\lambda)$ EA,  define its distance function as follows:
\begin{align}
d^{(1+\lambda)} (x)= \left\{
\begin{array}{lll}
0, &\text{if } x=s_0,\\
 g^{(1+1)}(s_{K+L+1}), &\text{if } x=s_{K+L+1},\\
\lambda g^{(1+1)}(x), &\text{otherwise}.
\end{array}
\right.
\end{align} 

Since $\Phi^{(1+\lambda)}_0=s_{K+L+1}$ and $f(s_{K+L}) > f(s_K)$, there are two potential events.
\begin{enumerate}
\item The $(1+\lambda)$ EA moves  from $ s_{K+L+1}   $ to $  s_{K} $. This event could happen if and only if all mutated children are $s_{K}$.  The probability  for the event happening is $p^{\lambda}$.

\item The $(1+\lambda)$ EA moves  from $  s_{K+L+1} $ to $ s_{K+L} $. This event could happen if and only if at least one mutated child is $s_{K+L}$.   The probability  for the event happening is $1-p^{\lambda}$.
\end{enumerate}

Calculate the pointwise drift at $ s_{K+L+1} $ as follows: 
\begin{align}
  &\Delta^{(1+\lambda)}( s_{K+L+1} )   \nonumber\\
=& (1-p^{\lambda} )( g^{(1+1)} (s_{K+L+1}) -\lambda g^{(1+1)}(s_{K+L}))  \nonumber\\
&+p^{\lambda}   (g^{(1+1)} (s_{K+L+1})- \lambda g^{(1+1)} (s_{K} ))\nonumber\\
= & (1-p^{\lambda} )[1+pK+(1-p)L -\lambda L]  \nonumber\\
&+p^{\lambda} [1+pK+(1-p)L-\lambda K] \nonumber \\
=& 1+pK +(1-p)L-\lambda L+p^{\lambda} \lambda L -p^{\lambda}\lambda K \nonumber\\
=&1+(p-p^{\lambda} \lambda) (K-L)+L(1-\lambda) \nonumber\\
 >&\lambda. \quad ( \text{use } \lambda<\lambda^* \text{ and } (\ref{equConditionTwoPathsII}))
\end{align}
Calculate the pointwise drift at $ s \in \{s_1, \cdots, s_{K+L} \}$, 
\begin{align*}
\Delta^{(1+\lambda)}( s  )&=\sum_{s' \in \mathcal{S}} [ \lambda g^{(1+1)} (s) -  \lambda g^{(1+1)} (s')]\\& = \Delta^{(1+1)}( s  )=\lambda. 
\end{align*}
 
Since $\Phi_0=s_{K+L+1}$, the average drift $\bar{\Delta}^{(1+\lambda)}_0  >\lambda$. For any $t\ge 1$, $\Phi^{(1+\lambda)}_t$ has left the point $s_{K+L+1}$, then   average drift
$ 
\bar{\Delta}^{(1+\lambda)}_t  =\lambda 
$. According to Theorem~\ref{thePSLowerBound},   we get that  $PS(1+\lambda \mid s_{K+L+1}) >\lambda$.
\end{IEEEproof}

Proposition~\ref{proTwoPathsII} states that using a  population (within the scalability threshold) may reduce  the expected running time on   two-paths-II functions within the scalability threshold. 
It is easy to understand the prosition.   There are two paths to the optimum: short and long. Using a lager population, the short path is more likely to be chosen than the long path. Thus the expected hitting time is reduced.

\begin{example}
\label{exaTwoPathsII}
Consider an instance of two-paths-II functions.  Let $x \in \{0,1\}^n$ and $|x|$ denote its number of 1-valued bits. 
\begin{equation}
\label{equTwoPathsII}
 f(x)=\left\{
\begin{array}{lll}
n, & \text{if } |x|=0 \text{ or } |x| = n,\\
-|x|, & \text{if } |x| \le \theta \text{ where }\theta=n-2,\\
|x|, & \text{otherwise}.
\end{array}
\right.
\end{equation}   
The $(1+\lambda)$ EA  is the same as that in Section~\ref{secTwoPathsI}. 
The $(1+1)$ EA can find an optimal solution in less than $n$ generations.  
   
\begin{itemize}
\item  \textbf{Adaptive mutation.}   
\begin{enumerate}
\item  When $t =0$,   flip one of  1-valued bits with the probability 0.5, otherwise flip one of 0-valued bits. In this way, generate $\lambda$ children as $\Psi_0$. 

\item When $t \ge 1$, if $f(\Phi_t) >f(\Phi_{t-1})$ and $\Phi_t$ is generated by flipping a 0-valued bit (or 1-valued) in $\Phi_{t-1}$,   flip   a 0-valued bit (or 1-value bit). Then generate $\lambda$ children as $\Psi_t$. 

\item  When $t \ge 1$, if $f(\Phi_t)= f(\Phi_{t-1})$,  either  flip one of  1-valued bits with probability 0.5 or  flip one of 0-valued bits.  Then generate $\lambda$ children as $\Psi_t$. 
\end{enumerate}

\item  \textbf{Elitist Selection.}   Select the best individual  from $\Phi_t \cup \Psi_t$. 

\end{itemize}    
 Table~\ref{tabTwoPathsII} shows that using a   population  reduces the expected running time. 
 
\begin{table}[htb]
\caption{Experimental results for Example~\ref{exaTwoPathsII} averaged over 1000 runs. $n=1000$. The   EA   starts at $x_0$ with  $|x_0|=998$.}
\label{tabTwoPathsII}
\centering
\begin{tabular}{c|c|c|c|c|c }  
    \toprule
    population size   & 1 &  5 & 9   &13 &17  \\
    \midrule
    hitting time  & 507& 28& 2& 2& 2 \\ 
    \midrule
    running time   & 507& 140& 18& 26& 34  \\ 
    \bottomrule
\end{tabular}
\end{table}

\end{example} 
 
Notice that the function in the  example  is a unimodal function under Hamming distance. The result shows that using a population shortens the expected running time of an EA on a unimodal function  in terms of the distance-based fitness landscape.

\subsection{Case study 6: deceptive-like functions}
\label{secCase5}
\begin{figure}[ht]
\centering
\begin{tikzpicture}
\begin{axis}[width=8cm,height=4.5cm,
xmax=7,ymax=14,
axis lines=left,
ticks=none, 
legend style={draw=none},
xlabel=$g^{(1+1)}(x)$, ylabel=$f(x)$,  
]  
\addplot+[nodes near coords,only marks,
point meta=explicit symbolic]
table[meta=label] {
x y label
0 10 $s_0$
1 1 $s_1$
2 6 $\,$
3 5 $\,$
4 2 $\,$
5 3 $s_{K-1}$
6 8 $s_K$
};
\end{axis}
\end{tikzpicture}

  \caption{A deceptive-like time-fitness landscape.   The $x$ axis is    the expected hitting time of the $(1+1)$ EA. The $y$ axis is the fitness function.   }
\label{figMultiModal}
\end{figure}
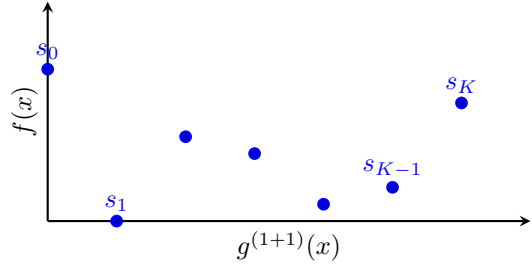

It is an intuitive  belief that using a population may shorten the runtime of EAs on   deceptive functions. This was  proven  for an elitist EA on Fully-Deceptive functions under Hamming distance~\cite{he2002individual}. In this case study, the conclusion is generalised  to deceptive-like functions in any finite set. Fully-deceptive functions and deceptive-like functions are defined as follows. 
\begin{definition}
Let $\mathcal{S}=\{ s_0, s_1, \cdots, s_K \}$. The fitness function satisfies  
$
f(s_0)>f(s_{K}) > \cdots >f(s_1).
$
  Given a $(1+1)$ elitist EA to maximise  $f$, $f$ is called  \emph{fully-deceptive-like}  to the $(1+1)$ EA if      $g^{(1+1)}(s_{K})>\cdots >g^{(1+1)}(s_1)$.   
\end{definition}
\begin{definition}
Let $\mathcal{S}=\{ s_0, s_1, \cdots, s_K \}$. The fitness function satisfies  
\begin{align}
\label{equDeceptiveLike}
f(s_0)>f(s_{K}) >\max \{f(s_1), \cdots, f(s_{K-1}) \}.
\end{align}
  Given a $(1+1)$ elitist EA to maximise  $f$,  $f$ is called  \emph{deceptive-like } to the $(1+1)$ EA if      $g^{(1+1)}(s_{K})>g^{(1+1)}(s_k)$ for any $k<K$. 
\end{definition}
 A deceptive-like time-fitness landscape is visualised in Fig.~\ref{figMultiModal}. 
 
Given a fitness function $f(x)$, consider an elitist $(1+1)$ uses global mutation for maximising $f(x)$. 

\begin{itemize}
\item    \textbf{Global Mutation.} For any $i$ and $j$, mutation   probability $P_m(s_i,s_j)>0$; For any $i < K$, mutation   probability $P_m(s_i, s_0) >  P_{\mathrm{m}}(s_K, s_0)$.

   \item  \textbf{Elitist Selection.} 
   select the best  from $\Phi_t \cup \Psi_t$.
   
\end{itemize}

The  fitness function $f$ is deceptive-like to the  $(1+1)$ EA because $g^{(1+1)}(s_{K})>g^{(1+1)}(s_i)$ for any $i<K$. This can be proven   by pointwise drift analysis. Let the distance function to be $d(s)=g^{(1+1)} (s_K)$ for $s \in \{ s_1, \cdots, s_K\}$ and $d(s)=0$ for $s=s_0$. The pointwise drift satisfies $\Delta^{(1+1)} (s)=1$ for $s=s_K$ and $\Delta^{(1+1)}(s_i) >1$ for $s_i \in \{s_1, \cdots, s_{K-1} \}$ due to   $P_m(s_i, s_0) >  P_{\mathrm{m}}(s_K, s_0)$. Hence $g^{(1+1)}(s_{K})>g^{(1+1)}(s_i)$.

Consider a $(\lambda+\lambda)$ EA (where $\lambda\ge 2$) using   global mutation   and  elitist selection with fitness diversity for maximising $f(x)$. 

\begin{itemize}
\item    \textbf{Global Mutation.} The same as that in the $(1+1)$ EA.

   \item  \textbf{Elitist Selection with Fitness Diversity.} 
\begin{enumerate}
\item     If the individuals in $\Phi_t \cup \Psi_t$ have the same fitness, then choose $\lambda$ individuals at random; 

\item   Otherwise,  first select one individual with the highest fitness  from $\Phi_t \cup \Psi_t$ and then select one individual with a lower higher  fitness.  Thereafter select $\lambda-2$ individuals from $\Phi_t \cup \Psi_t$ using any selection scheme. 

 \end{enumerate}   
\end{itemize}

We will show that the running time of the $(\lambda+\lambda)$ EA (where $\lambda \ge 2$) is shorter than that of the $(1+1)$ EA. First he following notation is introduced.   
\begin{align}
&\mathcal{S}_{\underline{K}}\defeq \{x \in \mathcal{S} ; f(x) < f(s_K) \}, \\
& \mathcal{P}_{\underline{K}}\defeq \{X \in \mathcal{P}; f(x) < f(s_K) \text{ for  any } x \in X \},\\
& P_{\mathrm{m}}(x, \mathcal{S}_{\underline{K}})\defeq \sum_{y \in \mathcal{S}_{\underline{K}}}   P_{\mathrm{m}}(x, y), \\
& P_{\mathrm{m}}(X, \mathcal{P}_{\underline{K}})\defeq \sum_{x \in X}   P_{\mathrm{m}}(x, \mathcal{S}_{\underline{K}}).
\end{align}

\begin{proposition}
\label{proDeceptive}
 Given  the $(1+1)$  elitist EA and $(\lambda+\lambda)$ EA (where $\lambda \ge 2$) for maximizing  any deceptive-like function,  if the  population size  satisfies $\lambda< \lambda^*$ where $\lambda^*$ is given by
  \begin{align}
 \label{equConditionCat3}
 \lambda^*=\frac{\scriptstyle   P_{\mathrm{m}}(s_K, \mathcal{S}_{\underline{K}})\min_{0<i<K}  P_{\mathrm{m}}(s_i, s_0)}{\scriptstyle (   P_{\mathrm{m}}(s_K, \mathcal{S}_{\underline{K}})+ \min_{0<i<k}  P_{\mathrm{m}}(s_i, s_0) ) P_{\mathrm{m}}(s_K, s_0)},
 \end{align} 
 then $PS(\lambda+\lambda \mid s_K) >\lambda$. The scalability threshold  is not les than $   \lambda^*$.
\end{proposition}

\begin{IEEEproof}
For the $(\lambda+\lambda)$ EA, given a population $X$, let $f(X) =\max \{f(x); x \in X\}$ the maximal fitness of its individuals. Define the distance distance as follows:  
\begin{align}
d(X) =\left\{
\begin{array}{llll}
 d_0=0, &\scriptstyle\text{if } f(X)=f(s_0),\\
  d_K=g^{(1)}(s_K), &\scriptstyle \text{if } f(X)=f(s_K),\\
d_{\underline{K}}= \scriptstyle \frac{\lambda}{ \min_{0<i<K}  P_{\mathrm{m}}(s_i,s_0)}, &\scriptstyle\text{if } f(X)<f(s_K).
\end{array}
\right.
\end{align} 

We calculate the pointwise drift at $X =(s_K, \cdots, s_K)$:
\begin{align}
&\Delta^{(\lambda+\lambda)}(s_K, \cdots, s_K) \nonumber \\
= &P(X, \mathcal{P}_\opt)(d_K-d_0) +P(X, \mathcal{P}_{\underline{K}})(d_K-d_{\underline{K}}) \nonumber \\
 >&P(X, \mathcal{P}_\opt)(d_K-d_{\underline{K}}) +P(X, \mathcal{P}_{\underline{K}})(d_K-d_{\underline{K}})\nonumber\\
=&\left( P(X, \mathcal{P}_\opt)  +P(X, \mathcal{P}_{\underline{K}})\right) (d_K-d_{\underline{K}})\nonumber\\
>&   P_{\mathrm{m}}(s_K, \mathcal{S}_{\underline{K}}) (d_K-d_{\underline{K}})\nonumber\\
=&   P_{\mathrm{m}} (s_K, \mathcal{S}_{\underline{K}}) \left (\frac{1}{   P_{\mathrm{m}}(s_K, s_0)}-  \frac{\lambda}{\min_{0<i<K} P_{\mathrm{m}}(s_i,s_0)} \right) \nonumber\\
>& \lambda. \qquad (\text{use }  \lambda < \lambda^* \text{ and } (\ref{equConditionCat3})) \label{equDriftCat3Pointwise1}
\end{align}
Calculate the pointwise drift at any $X   \in \mathcal{P}_{\non}$ in which at least one of its individuals is not ${s}_K$:
\begin{align}
\Delta^{(\lambda+\lambda)}(X) =& P(X, \mathcal{P}_\opt)(d_{\underline{K}}-d_0)  \nonumber \\
\ge&    \frac{\lambda \max_{x  \in X}  P_{\mathrm{m}}(x, s_0) }{\min_{0<i<k}  P_{\mathrm{m}}(s_i,s_0)} \ge \lambda. \label{equDriftCat3Pointwise2}
 \end{align}
 
 Since $\Phi^{(\lambda+\lambda)}_0=(s_K, \cdots, s_K)$,  from (\ref{equDriftCat3Pointwise1}) we have the average drift   
 $\bar{\Delta}^{(\lambda+\lambda)}_0  >\lambda$. For any $t \ge 1$,   we know,  $\bar{\Delta}^{(\lambda+\lambda)}_t \ge \lambda$. 
  According to Theorem~\ref{thePSLowerBound},  we have $PS(\lambda+\lambda \mid s_K) >\lambda$.
\end{IEEEproof}

Here is an explanation of the proposition. In the $(1+1)$ algorithm, elitist selection cannot accept an worse solution. But in the population-based EA, selection with the fitness diversity can accept a worse solution. This helps the EA.

\begin{example}
\label{exaDeceptive}
Consider an instance of Fully-Deceptive functions.    Let $x \in \{0,1\}^n$ and $|x|$ denote its number of 1-valued bits. 
\begin{equation}
\label{equFullyDec}
 f(x)=\left \{
\begin{array}{llll}
  n, & \text{if }  |x| =0 \text{ or } |x| =n,\\
\min \{ |x|, n-|x|\} , &\text{otherwise}.  
\end{array}
\right.
\end{equation} 
There are two optima: $0\cdots0$ and $1\cdots 1$. Consider  a $(\lambda+\lambda)$  EA  using elitist selection and bitwise mutation:
\begin{itemize}
\item    \textbf{Bitwise Mutation.}   Flip each bit with probability $1/n$. Each  parent generates  one child.

   \item  \textbf{Elitist Selection + Random Selection.} 
     Select one individual with the highest fitness  from $\Phi_t \cup \Psi_t$ and then  select $\lambda-1$ individuals from $\Phi_t \cup \Psi_t$ at random. 
 
\end{itemize}

Table~\ref{tabDecptive} shows that using a  population  reduced the expected running time.
\begin{table}[htb]
\caption{Experimental results for Example~\ref{exaDeceptive} averaged over 1000 runs. $n=10$. The EA starts at $x_0$ with  $|x_0| =5$.}
\label{tabDecptive}
\centering
\begin{tabular}{c|c|c|c|c|c }  
    \toprule
    population size   & 1 &  5 & 9   &13 &17  \\
    \midrule
    hitting time  & 82774& 1411& 370& 188& 116 \\ 
    \midrule
    running time   & 82774& 7055& 3330& 2444& 1972 \\ 
    \bottomrule
\end{tabular}
\end{table}

\end{example}

\section{Discussion (Case study 7) }
\label{secDoscussion}
It must be pointed out that population scalability depends on the benchmark EA. Let's show this through a simple instance of Fully-Deceptive functions.
Let $x \in \{ 0,1\}^2$ be a binary string with length 2. The fitness function  is given by
\begin{align}
f(x)=
\left\{
\begin{array}{llll}
3, &\mbox{if } |x|=0,\\
|x|, &\mbox{if } |x|=1,2.
\end{array}
\right.
\end{align}

If   the  benchmark $(1+1)$ EA is changed from elitist selection to random selection, then using a population cannot shorten the running time any more. 
\begin{itemize}
\item    \textbf{Bitwise Mutation.}   Flip each bit with probability $1/n$.

   \item  \textbf{Random Selection.} 
     Select one individual  from $\Phi_t \cup \Psi_t$ at random. 
      
\end{itemize}

Let $0,1,2$ represent  the states  $|x|=0,1,2$ respectively. Transition probabilities of the $(1+1)$ EA among non-optimal states are given in Table~\ref{tabProbabiltiesI}.
\begin{table}[ht]
\caption{Transition probabilities among non-optimal states}
\label{tabProbabiltiesI}
\centering
\begin{tabular}{c|cccc}
   \toprule
$p(x,y)$ & $ 1$ & $ 2$ \\
\midrule
$ 1$ & $\frac{5}{8}$ & $\frac{1}{8}$  \\
\midrule
$ 2$ &  $\frac{1}{4}$ &   $\frac{2}{4}$\\
\bottomrule
\end{tabular}
\end{table}

According to Theorem~\ref{theFundamentalMatrix}, the expected hitting time  of the $(1+1)$ EA satisfies a linear equation  system:
\begin{align}
\left\{
\begin{array}{llll}
\frac{3}{8} g^{(1+1)}(1)-\frac{1}{8} g^{(1+1)}(2)=1,\\
- \frac{1}{4} g^{(1+1)}(1)+\frac{2}{4} g^{(1+1)}(2)=1.
\end{array}
\right.
\end{align}

Solving the equations, we get the expected hitting time  as follows:  
\begin{align}
g^{(1+1)}(1)= g^{(1+1)}(2)=4. 
\end{align}
Starting from any non-optimal state, the expected running time of the $(1+1)$ EA equals to its expected hitting time ($=4$).

Now we consider a simple $(2+2)$ EA which runs the two copies of the above $(1+1)$ EA independently. Let $(i,j)$ represent  the population state such that $|x_1|=i,$ $|x_2|=j$ where $i,j=0,1,2$. Transition probabilities of the (2+2) EA among non-optimal states are given in Table~\ref{tabProbabiltiesII}.
\begin{table}[ht]
\caption{Transition probabilities among non-optimal states}
\label{tabProbabiltiesII}
\centering
\begin{tabular}{c|cccc}
   \toprule
$p(x,y)$ & $ (1,1)$ & $(1, 2)$ &$(2,1)$ & $(2,2)$ \\
\midrule
$ (1,1)$ & $\frac{5}{8} \cdot \frac{5}{8}$ & $\frac{5}{8} \cdot \frac{1}{8}$ & $\frac{1}{8} \cdot \frac{5}{8}$ & $\frac{1}{8} \cdot \frac{1}{8}$ \\
\midrule$ (1,2)$ & $\frac{5}{8} \cdot \frac{1}{4}$ & $\frac{5}{8} \cdot \frac{2}{4}$ & $\frac{1}{8} \cdot \frac{1}{4}$ & $\frac{1}{8} \cdot \frac{2}{4}$ \\
\midrule$ (2,1)$ & $\frac{1}{4} \cdot \frac{5}{8}$ & $\frac{1}{4} \cdot \frac{1}{8}$ & $\frac{2}{4} \cdot \frac{5}{8}$ & $\frac{2}{4} \cdot \frac{1}{8}$ \\
\midrule$ (2,2)$ & $\frac{1}{4} \cdot \frac{1}{4}$ & $\frac{1}{4} \cdot \frac{2}{4}$ & $\frac{2}{4} \cdot \frac{1}{4}$ & $\frac{2}{4} \cdot \frac{2}{4}$ \\
\bottomrule
\end{tabular}
\end{table}

According to Theorem~\ref{theFundamentalMatrix}, the expected hitting time  of the (2+2) EA satisfies a linear equation  system:
\begin{align}
\left\{
\begin{array}{llll}
 \frac{39}{64} g^{(2+2)}(1,1)-\frac{5}{64} g^{(2+2)}(1,2)\\
 \qquad -\frac{5}{64} g^{(2+2)}(2,1) -\frac{1}{64} g^{(2+2)}(2,2) =1,\\
 -\frac{5}{32} g^{(2+2)}(1,1)+\frac{22}{32} g^{(2+2)}(1,2)\\
 \qquad -\frac{1}{32} g^{(2+2)}(2,1) -\frac{2}{32} g^{(2+2)}(2,2) =1,\\
 -\frac{5}{32} g^{(2+2)}(1,1)-\frac{1}{32} g^{(2+2)}(1,2)\\
 \qquad +\frac{22}{32} g^{(2+2)}(2,1) -\frac{2}{32} g^{(2+2)}(2,2) =1,\\
 -\frac{1}{16} g^{(2+2)}(1,1)-\frac{2}{16} g^{(2+2)}(1,2)\\
 \qquad -\frac{2}{16} g^{(2+2)}(2,1) +\frac{12}{16} g^{(2+2)}(2,2) =1.
\end{array}
\right.
\end{align}

Solving the equations, we get the expected hitting time  as follows:  
\begin{align}\scriptstyle
 g^{(2+2)}(1,1)= g^{(2+2)}(1,2) 
= g^{(2+2)}(2,1) = g^{(2+2)}(2,2) \displaystyle =\frac{16}{7}.
\end{align}
Staring from any non-optimal state, the expected running time   of the $(2+2)$ EA is $ 32/7 \ge 4$. It is greater than that of the $(1+1)$ EA. 

Therefore using a population doesn't shorten the expected running time on the Fully-Deceptive function if the EA uses random selection. The reason is simple: the $(1+1)$ EA with random selection can accept a worse solution. Hence using a population doesn't help too much. 

\section{Conclusions}
\label{secConclusions}
This paper proposes  population scalability for studying how population-size affects the computation time of population-based EAs.  Population scalability is the ratio of the expected hitting time  between a benchmark EA and an EA using a larger population size.   Average drift analysis is presented as a tool of comparing the expected hitting time  of two EAs and estimating lower and upper bounds on population scalability.  


The main results can be regarded as a rigorous analysis of several intuitive  beliefs. 
\begin{enumerate}
\item  ``Using a population may reduce the expected hitting time of an EA to find an optimal point.''  This belief is not always true. Two counter-examples are given, which are  a $(1+\lambda)$ EA  on two-paths-II functions and a $(\mu+1)$ EA on unimodal functions.

\item ``Using a population cannot shorten the expected running time of an elitist EA on unimodal functions.'' This belief is always true  for any $(\mu+\lambda)$ EAs with elitist selection on   unimodal functions in terms of the time-fitness landscape. But it is not always true in terms of the  distance-based fitness landscape.

\item ``Using a population can reduce the expected running time of an EA on  fully-deceptive functions.'' This belief is not always  true. It is true for a $(\lambda+\lambda)$  EA  if the benchmark $(1+1)$ EA uses elitist selection, but not  true if the benchmark EA uses random selection.  
\end{enumerate}

There exist many open questions, for example,  how to estimate the scalability threshold? how to analyse population scalability of EAs with crossover? what practical criteria can be used in judging population scalability? what is the optimal population size that maximises population scalability?

\section*{Acknowledgement}    The authors would like to thank Dr. Tianshi Chen for his useful comments on deceptive functions and the late Dr. Boris Mitavskiy for his useful comments on two-paths functions.  

 \end{document}